\documentclass{article}
\usepackage{xspace}
\usepackage[final]{colm2026_conference}

\usepackage{microtype}
\usepackage{hyperref}
\usepackage{url}
\usepackage{booktabs}

\usepackage{lineno}

\definecolor{darkblue}{rgb}{0, 0, 0.5}
\hypersetup{colorlinks=true, citecolor=darkblue, linkcolor=darkblue, urlcolor=darkblue}

\usepackage{latexsym}

\usepackage[T1]{fontenc}
\usepackage[utf8]{inputenc}

\usepackage{inconsolata}

\usepackage{amsmath}
\usepackage{amssymb}

\usepackage{multirow}   
\usepackage{array}      

\usepackage{graphicx}

\usepackage{subcaption}
\usepackage{longtable}
\usepackage{wrapfig}
\usepackage{listings}
\usepackage{enumitem}

\usepackage{xcolor}

\newcommand{\filleddiamond}{%
  \raisebox{0.15ex}{\rotatebox{45}{\rule{0.55em}{0.55em}}}%
}

\usepackage{tabularray}
\UseTblrLibrary{booktabs}

\definecolor{Mint}{RGB}{100,200,150}

\definecolor{orange}{RGB}{235, 158, 52}

\newcommand{\dataset}{\textsc{GAPA}\xspace}

\title{How Humans and LLMs Read Gender into ``Gender-Neutral'' Physical Descriptions}

\author{Yingjia Wan$^{1*}$ \And
  Lin L. Lin$^{1}$\thanks{Equal contribution.} \And
  Elisa Kreiss$^{1,2}$ \AND
  \\[-5ex] $^1$Department of Communication\\$^2$Department of Computer Science\\
  University of California, Los Angeles (UCLA)\\
  \texttt{\{alisawan, llin, ekreiss\}@ucla.edu}
  }

\begin{document}

\ifcolmsubmission
\linenumbers
\fi

\maketitle

\begin{abstract}

When foundation models describe people, recent work in AI fairness, accessibility, and ethics recommends avoiding inferred identity labels (e.g., ``she'', ``his'') in favor of seemingly ``objective'' physical descriptions (e.g., ``short hair'', ``a defined jawline''). Yet whether such descriptive language achieves gender-neutral communication remains an open empirical question. To study this, we introduce \textbf{\dataset~(Gender Associations of Physical Attributes)}, a dataset of 316 common physical attributes drawn from diverse sources and domains, paired with 14,706 gender-association ratings from 304 US-based annotators. Results of human ratings show that physical descriptions carry structured and graded gender associations among readers: 53\% of attributes are significantly more strongly associated with one gender than the others, with more consistent and distinctive associations for women and men than for non-binary identities. Next, we evaluate 16 LLMs across model families, sizes, and post-training variants against human ratings. The models partially recover human associations but exhibit systematic alignment biases, including compressed rating distributions, weaker alignment for associations with men, and asymmetric abstention that disproportionately targets the non-binary category. This abstention pattern is especially pronounced among instruction-tuned and proprietary models. 
Finally, we release the best-performing proxy model trained to predict humans' gender associations of descriptive language at scale and demonstrate its utility through a sociolinguistic analysis of character descriptions in LitBank.
Together, our findings provide the first empirical evidence that seemingly ``objective'' physical descriptions can retain systematic gender associations in human interpretation, and uncover systematic patterns of model--human misalignment. This challenges the assumption that replacing explicit gender labels with physical descriptions necessarily yields gender-neutral communication, and highlights downstream challenges in using such descriptions to communicate subjective identity categories in human--AI interaction.
\footnote{The dataset and code are available at ~\url{https://github.com/Yingjia-Wan/GAPA}, and the predictor model released at \url{https://huggingface.co/alisa-yingjia-wan/gapa-predictor-olmo2-7b}.}
\end{abstract}

\section{Introduction}


How should models describe a person? This question has become increasingly consequential as foundation models are deployed at scale in image-captioning systems, accessibility tools, content moderation pipelines, and assistive technologies. A range of research communities (e.g., fairness in machine learning, accessibility, critical AI studies, and human-computer interaction) have converged on a partial answer: when describing a person whose identity cannot be confirmed, models should avoid inferring categorical identity labels (race, gender, disability) and instead describe observable physical attributes \citep{bennett2021s,scheuerman2019computers,scheuerman2020we,hamidi2018gender}. Such ideology is influential in shaping both human image-description practice and model behavior, and recent foundation models increasingly default to descriptive rather than identity-inferring language when describing people \citep{marin2026vision,hanley2021computer}. The reasoning behind this looks epistemic and ethical: describing what is visually present, rather than what is inferred about the person, transfers the inference back to the reader and supposedly avoids imposing identity claims the system cannot verify.

However, such reasoning rests on an assumption that has not been empirically tested: that descriptive language is meaningfully gender-neutral. The twist here is: if ``a chiseled jawline'' or ``a defined waist'' carries gender associations as systematic as those carried by ``a man'' or ``a woman'', then the strategy may shift gendered communication into a less visible register rather than reducing it. This raises some critical, underexplored questions: do isolated physical descriptors, stripped of visual context, carry gender associations among human readers? If so, how strong, how consistent across people, and how distributed across types of attribute? Furthermore, do language models reproduce these associations, amplify them, or structurally distort them?


\begin{figure*}[t]
    \centering

    \includegraphics[width=\textwidth]{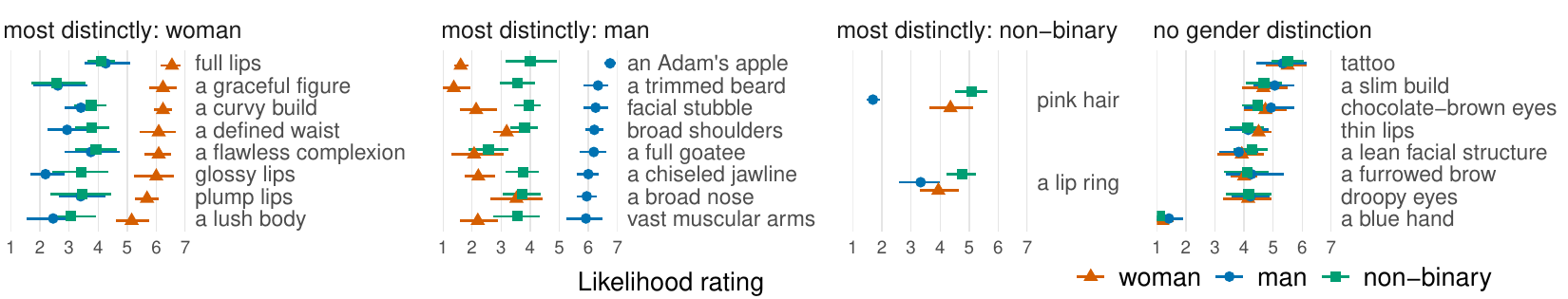}
    \vspace{-15pt}
    \caption{An excerpt of the most gender-distinctive attributes with significant gender differences ($p < 0.05$) in each ranking pattern, with their per-gender association ratings on the y-axis.
    See \autoref{fig:attribute_results} in the Appendix for the complete rating plots of all attributes with significant gender differences in the six ranking patterns.}
    \vspace{-15pt}
    \label{fig:top_attributes}
\end{figure*}

It remains a significant blind spot to what extent linguistic descriptions of physical characteristics themselves are gendered,
with little evidence in prior work across disciplines. Sociolinguistic and psychological research has long established that humans infer gender from visual and acoustic cues within hundreds of milliseconds \citep{besson2017face,freeman2016more, brown1993gives,freeman2012social} and that language carries graded, context-sensitive gendered meanings \citep{eckert2003language}. In NLP, extensive work on gender bias has developed diagnostic benchmarks \citep{zhao2018gender,bolukbasi2016man,caliskan2017semantics} and mitigation methods \citep{dong2024disclosure,zhao2019gender}. These benchmarks typically evaluate whether foundational models associate gender with occupations, traits, pronouns, etc. However, they do not examine whether the descriptive vocabulary itself carries systematic gender associations. This gap is especially consequential because current practices that replace explicit identity labels with ostensibly neutral physical descriptions implicitly assume that such descriptions do not themselves transmit gendered information.




To address this gap, this study challenges the assumption that physical attribute descriptions function as semantically gender-neutral fallbacks, by presenting the first series of evidence from human and LLM experiments. We make the following contributions:

\begin{enumerate}[leftmargin=1.5em]
\vspace{-5pt}
\item \textbf{Empirical evidence that physical attributes carry systematic gender associations.} We introduce GAPA (Gender Associations of Physical Attributes), a \textbf{human-annotated dataset} of 316 physical attribute descriptions sourced from three complementary channels (LLM generation, human elicitation, and contemporary fiction) annotated with 14,706 ratings from 304 U.S.-based participants on perceived associations with women, men, and non-binary gender categories (see~\autoref{fig:top_attributes}). 
While much prior work has focused on quantifying visual gender associations, GAPA allows us to do this in the linguistic modality, providing a crucial foundation for empirical studies on ``gender-neutral'' descriptions of people's appearances.
Section~\ref{sec:dataset} introduces the construction process and statistics of the dataset, while Section~\ref{sec:human_exp} reports the human experiments from which we collect the human annotations of how physical attributes from \dataset are associated with each gender category.

\item \textbf{Systematic Evaluation of LLM Gender Representations.} After establishing that humans carry strong gender associations with physical attributes, it becomes imperative to understand to what extent models faithfully encode or distort them in meaningful ways, as this alignment marks the basis for human--AI interaction downstream. We evaluate 16 state-of-the-art LLMs on GAPA against human ratings, and find that models only partially recover the human pattern. This highlights that LLMs, too, carry gender associations with linguistic descriptors of physical attributes, yet exhibit systematic distortions on multiple levels. See Section~\ref{sec:model_exp} for detailed results and discussions.


\item \textbf{Studying Gender Associations at Scale.} Our findings also provide a new opportunity for studying linguistic gender representations at scale. To that end, we develop and release a proxy predictor of human gender associations for automatically quantifying how humans associate gender with descriptive language. Our selected predictor, fine-tuned on \dataset, achieves r=0.76 against held-out human ratings after extensive hyperparameter search and model ablations. We demonstrate its utility via a sociolinguistic analysis of character descriptions in LitBank \citep{litbank_bamman2019annotated} (Section~\ref{sec:scaleup}).

\end{enumerate}

\vspace{-5pt}
\section{Related Work}
\vspace{-5pt}

Linguistic and social science research conceptualizes gender not as a fixed category but as a socially constructed, context-dependent phenomenon \citep{west1987doing, eckert2003language}. From this perspective, gender is indexed through speech patterns and styles, making it a graded dimension rather than a binary one \citep{bucholtz2005identity}. Empirical work demonstrates that humans infer gender from a wide array of cues—including voice, facial morphology, and motion \citep{johnson2005perceiving}—and that these associations are systematic and context-sensitive.

In NLP, extensive work has examined the explicit and implicit gender biases that language models encode, producing influential diagnostic benchmarks like WinoBias and developing mitigation strategies such as data balancing and fine-tuning \citep{bolukbasi2016man, zhao2018gender, dong2024disclosure}. More recent research challenges the predominantly binary treatment of gender, revealing that models struggle with gender-neutral pronouns, show lower performance on non-binary categories, and are inconsistent in generating gender-neutral language \citep{hossain2023misgendered, you2024beyond, savoldi2025mind}.

However, this body of work has almost exclusively focused on biases associated with labels (e.g., occupations, pronouns) rather than on the descriptive vocabulary itself (but see \cite{gao2025measuring} for a first exploration of physical attributes as stimuli). It remains underexplored whether seemingly objective physical descriptions, increasingly recommended as a neutral alternative to identity labels \citep{bennett2021s}, carry systematic gender associations. Our work addresses this gap by investigating the gendered nature of physical description language and how it is represented by both humans and language models.

\vspace{-0.5em}
\section{\dataset: A Collection of Physical Attributes}
\label{sec:dataset}
\vspace{-0.5em}

\begin{wraptable}{r}{0.6\textwidth} \vspace{-10pt} \centering \scriptsize
\setlength{\tabcolsep}{2.5pt} \renewcommand{\arraystretch}{1.1} \begin{tabular}{lrrrr} \toprule \dataset  Set & Attributes & Participants & \shortstack[l]{Questions} & Total Ratings \\ \midrule LLM-generated & 170 & 178 & 510 & 8,406 \\ Novel-extracted & 96 & 80 & 288 & 2,300 \\ Human-written & 50 & 46 & 150 & 4,000 \\ \midrule Overall & 315 & 304 & 945 & 14,706 \\ \bottomrule \end{tabular} \vspace{-5pt} \caption{Summary statistics of \dataset.} 
\vspace{-10pt}
\end{wraptable}

In order to systematically investigate how humans and language models associate gender with text descriptions of physical attributes, we built \dataset, the largest existing collection of linguistic descriptions of physical attributes. Building a lexicon of physical attributes serves two purposes: (1) to provide a structured resource for analyzing variation in appearance-based language across sources, and (2) to enable evaluation of whether physical attributes are systematically perceived as gendered.

To facilitate diversity of physical attributes, we collected them from  distinctive sources based on LLM-generated attributes (providing broad categorical coverage; n=170), human-written attributes (ensuring high-frequency terms; n=50), and attributes extracted from six novels (covering more unusual and stylized references; n=96), resulting in a comprehensive collection of 316 physical attributes. We provide all procedural specifications on the selection process in \autoref{app:attrselection}.



\section{Physical Attributes Carry Systematic Gender Associations for Humans}
\label{sec:human_exp}

After compiling the physical attribute lexicon, we conducted a human-subject experiment to measure how strongly participants associated each attribute with different gender categories: \textit{woman}, \textit{man}, and \textit{non-binary person}\footnote{We include \textit{woman}, \textit{man}, and \textit{non-binary person} because they reflect commonly recognized gender categories in contemporary U.S. English. These categories are not intended to represent the full range of gender identities, nor to imply that gender is naturally discrete. Instead, they serve as reference points for examining how physical attributes are associated with gender in everyday language \textit{beyond a simple woman--man contrast}.}. Specifically, we investigate: (1) whether there is an overall systematic difference in gender association ratings in \dataset, averaged across attributes and participants; (2) whether there exists a consistent shared association of gender for each attribute, alongside heterogeneity in how gender differences manifest at the attribute level in terms of direction and magnitude.

\vspace{-0.5em}
\subsection{Experiment Details}
\vspace{-0.5em}


The annotated dataset of \dataset~includes 304 participants (152 female; ages 18–77, $M = 41.36$) who passed at least four of five attention checks. Analyses were conducted controlling for participant demographics (e.g., gender, age) and attribute source. Full details of human participant demographics and the recruitment process are provided in Appendix \ref{app:participant-demographics}. 

On each trial, participants used a 7-point Likert scale to answer: \textit{``How likely is it for someone to say that a {woman/man/non-binary person} has {a given physical attribute}?''} Asking what \textit{someone} would say captures culturally available associations rather than participants’ explicit endorsement of those associations, as individuals may possess stereotype knowledge that is different from personal belief \citep{devine1989stereotypes, kunda2003when}.

Using a cross-classified design, participants rated randomly sampled attributes. To minimize carryover effects and direct cross-gender comparisons, each participant rated a given attribute for only one target gender. Counterbalancing ensured an even distribution of target gender and trial order. Each participant completed 55 trials, yielding an average of 17 ratings per attribute, per gender category.

\vspace{-0.5em}
\subsection{Metrics}
\vspace{-0.5em}

We define two complementary metrics to quantify the gender association of physical attributes. \textbf{Gender Association} is the mean rating for each attribute–gender pair, capturing how strongly participants associate a physical attribute with a given gender category. \textbf{Gender Distinctiveness} is the difference between the highest and second-highest mean ratings across the three gender categories for a given attribute, capturing how clearly that attribute differentiates one category from the others.


Together, these measures distinguish attributes that are strongly associated with a gender category from those that are also clearly differentiated across categories. This approach is consistent with prior work showing that category membership and social judgments are represented in graded rather than binary terms \citep{rosch1975cognitive, rosch1975family, barsalou1985ideals}, and that person perception depends not only on category activation but also on how diagnostic cues are for distinguishing among categories \citep{freeman2011dynamic, oosterhof2008functional}.

\vspace{-0.5em}
\subsection{Results}
\label{sec:human_exp_results}
\vspace{-0.5em}

\paragraph{Dataset-Level Gender Effects}

Across the dataset, attributes were associated more with women than men ($\beta = 0.26$, $SE = 0.04$, $z = 6.77$, $p < .001$), and least with non-binary people ($\beta = 0.43$, $SE = 0.04$, $z = 11.68$, $p < .001$), providing first evidence that physical attributes show structured gender profiles and are not perceived as generally gender-neutral. There are two potential explanations for the significant distinctions between genders: the dataset may overrepresent attributes that are culturally coded as feminine, or the pattern may reflect broader biases in language use, where physical descriptors are unevenly distributed across gender categories. Ultimately, the data clearly reveals that people have nuanced gender associations with physical attributes that we now aim to unpack.

\vspace{-1em}
\paragraph{Attribute-Level Gender Effects}

\begin{wrapfigure}{r}{0.57\textwidth}
    \centering
    \vspace{-8pt}
    \includegraphics[width=0.55\textwidth]{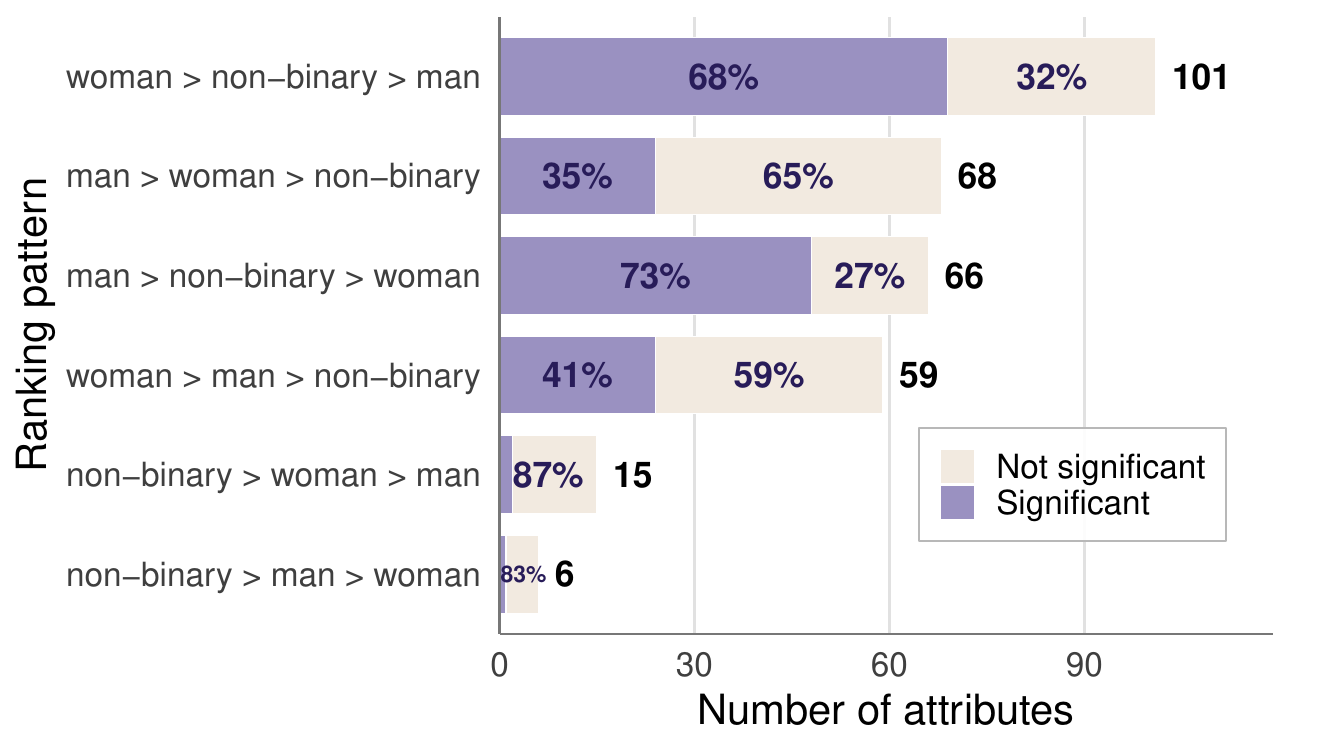}
    \vspace{-8pt}
    \caption{Attribute counts for gender-ranking patterns and their significance percentages ($p < 0.05$).}
    \label{fig:rq3_count_plot}
    \vspace{-20pt}
\end{wrapfigure}

Since different attributes may be associated with gender categories to varying extent, we also conduct a significance test for each attribute in \dataset, to examine heterogeneity in gender effects across attributes.

Over half of the attributes (53\%) are significantly more associated with one particular gender than other genders. Importantly, the direction and strength of attribute-level gender differences vary systematically across ranking patterns (\autoref{fig:rq3_count_plot}). Attributes that align with conventional binary gender distinctions (e.g., \textit{woman $>$ non-binary $>$ man} and \textit{man $>$ non-binary $>$ woman}) are both more frequent and substantially more likely to yield significant effects (68\%–73\%). In contrast, attributes where non-binary individuals are rated highest are the rarest and exhibit much lower rates of significance. \autoref{fig:top_attributes} provides the qualitative case studies for the attributes in each ranking pattern of gender ratings. 


Overall, the \textit{attribute-level heterogeneity} suggests that gender associations carried in physical attributes are structured around a binary axis, with non-binary categories occupying a less strongly encoded region of the semantic space. In this sense, ranking patterns can be interpreted as approximations of culturally shared notions of ``femininity'' and ``masculinity'', where attributes strongly aligned with these poles produce clearer and more detectable statistical effects. This intuition is supported by a significant negative correlation of man vs. woman attribute associations ($r = -.16$, $p = .004$).



\begin{figure*}
    \centering
    \includegraphics[width=\linewidth]{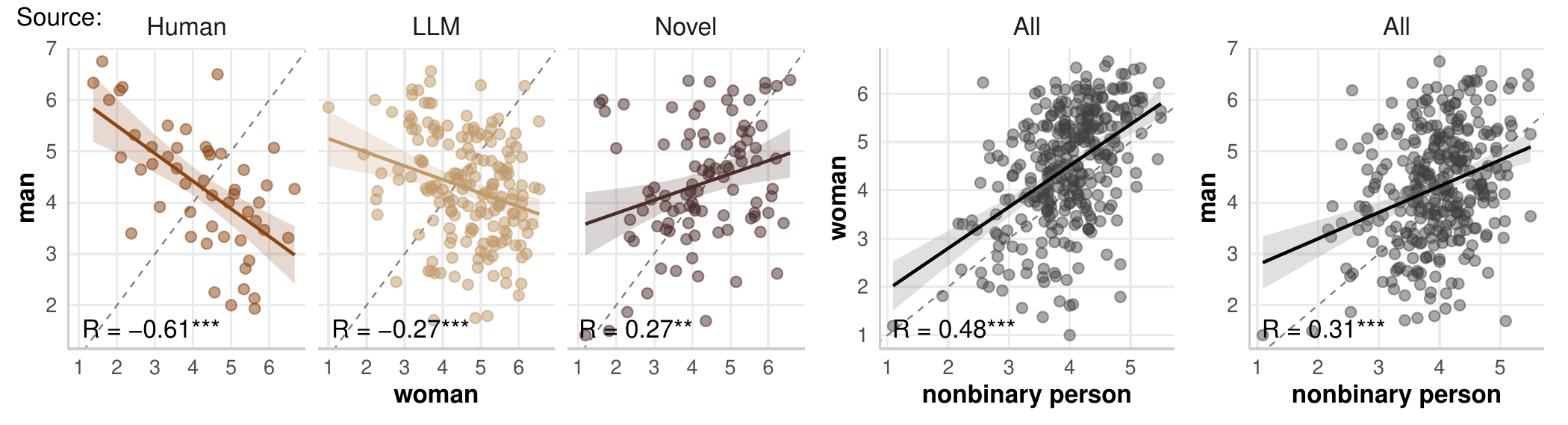}
    \caption{Correlations among mean gender-association ratings across physical attributes. Each point represents an attribute ($p < 0.05$). Man--Woman correlations are further split by source domain of the physical attributes, showing significant variation across attributes.}
    \label{fig:corr}
    \vspace{-15pt}
\end{figure*}

In contrast, ratings for non-binary gender associations with both women and men were significantly positively correlated (see \autoref{fig:corr}), highlighting how even when attributes are judged as common for a gender, it doesn't mean they are gender-diagnostic. Interestingly, splitting up the man--woman correlations by how we sourced attributes shows significant variation. While the correlation is highly negative for the human-sourced examples ($r=-0.61$), this reverses for novel-extracted physical attributes ($r=0.27$). This variation highlights the importance of employing diverse sampling techniques and is the reason why replications across source domains are especially challenging but meaningful. 

Together, these correlations indicate that gender associations in physical description are structured rather than random. This pattern suggests that physical attributes are not simply distributed along a single binary opposition, but instead occupy partially shared and partially distinct regions of gendered meaning. In particular, non-binary associations appear to overlap more with attributes linked to women than with those strongly opposed to them.

\begin{wrapfigure}{r}{0.54\textwidth}
    \centering
    \vspace{-10pt}
    \includegraphics[width=\linewidth]{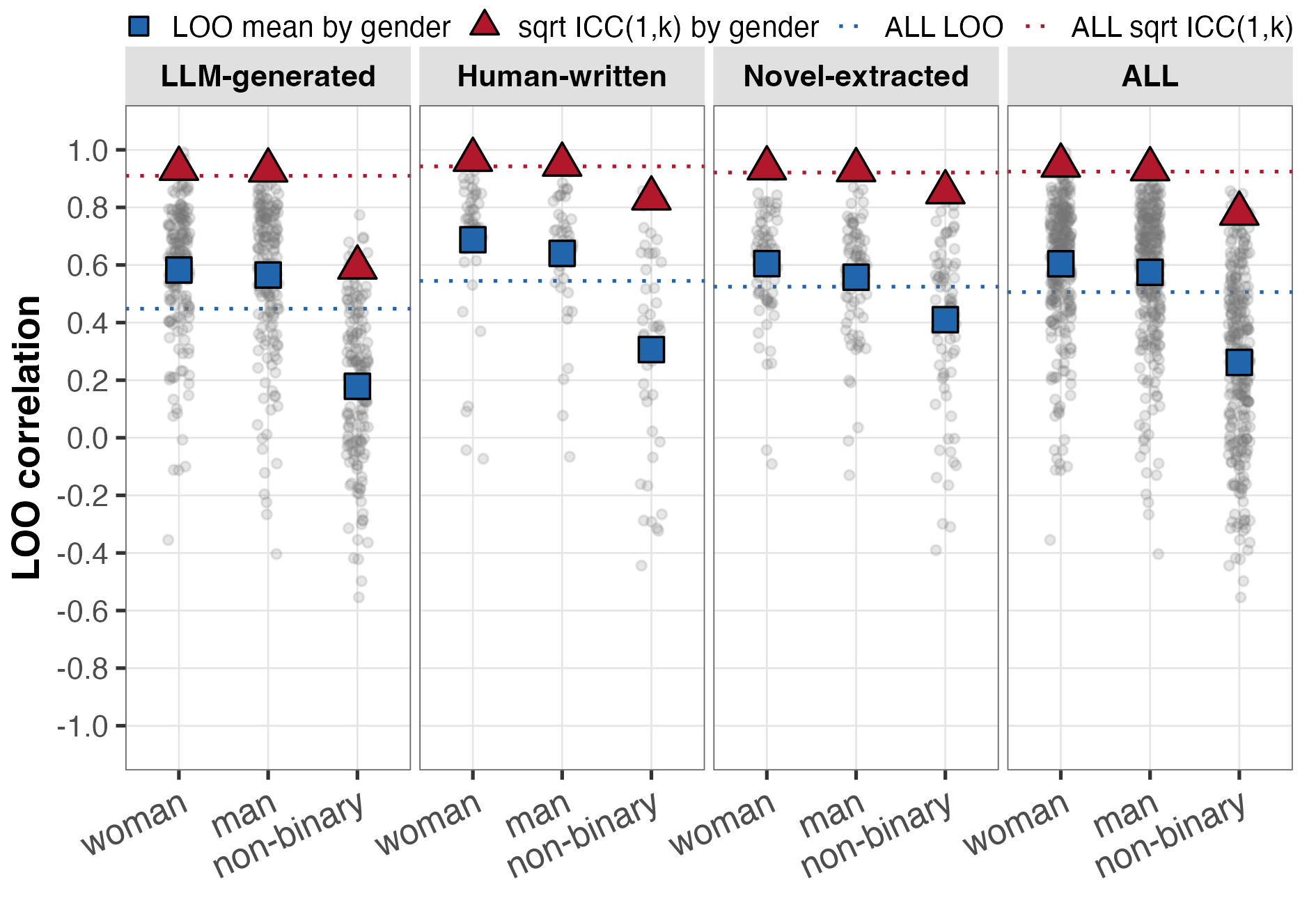}
    \caption{Human rater reliability in \dataset across three attribute sources, measured by $r^\text{LOO}$ and $\sqrt{\mathrm{ICC}(1,k)}$. The scatterplot shows $r^{\text{LOO}}$ per rater.}
    \vspace{-10pt}
    \label{fig:noise-ceiling}
\end{wrapfigure}

\paragraph{Human Uncertainty}
\label{sec:human_uncertainty}
Interpreting the gender association of physical attributes is inherently subjective and diverse among participants. To characterize this variability, we quantify \emph{human uncertainty} using two complementary reliability statistics: a leave-one-rater-out (LOO) correlation and the intraclass correlation coefficient (ICC). The formal calculation and scientific interpretation of the two measures for human uncertainty and LLM correlation baselines are elaborated on in Appendix \ref{app:noise-ceiling}.


In short, LOO measures how well a typical individual human rater agrees with the consensus by correlating each rater’s ratings with the mean ratings of all remaining raters across items, and then averaging across raters. This reflects the correlation performance of a \emph{typical} human annotator relative to the group. In contrast, ICC$(1,k)$ quantifies the reliability of the \emph{aggregated} human ratings by decomposing variance into signal (between-item variability) and noise (within-item, across-rater variability), estimating how much noise remains after averaging across raters.

\autoref{fig:noise-ceiling} shows the noise level of human ratings across attribute sources. Human rater reliability (measured by LOO and ICC) is consistently the highest for women, intermediate for men, and lowest for non-binary targets, indicating that ratings for non-binary attributes are less consistent across raters.

\section{Physical Attributes Carry Systematic Gender Associations for LLMs}
\label{sec:model_exp}
\vspace{-0.5em}

After establishing that humans have reliable gender associations with text descriptions of physical attributes, we now turn to compare these to LLM gender associations. 
We conduct a zero-shot evaluation similar to the human experiment, across a diverse set of 16 state-of-the-art language models, varying across open-weight vs. proprietary and instruct vs. base versions.

\vspace{-0.5em}
\subsection{Method}
\vspace{-0.5em}

For each physical attribute in our test set, we prompt the LLMs the same question as in the human subject experiment, with minor modifications to instruct the model to directly respond with a single integer rating on a 1--7 Likert scale (see full prompt in Appendix \ref{app:model_exp_prompt}).

Two complementary strategies to elicit ratings were explored: (1) direct generation: we directly extract the verbalized ratings in the model-generated response; (2) top-logit token extraction: we examine the model's output distribution at the first generated token position. Two methods showed consistency in results, as illustrated in Appendix~\ref{app:model_exp_method} in detail. We report the results from method (1).

\vspace{-0.5em}
\subsection{Evaluation Metric \& Baselines}
\vspace{-0.5em}

To assess how well a model's predictions align with the averaged human ratings across attributes, we adopt the Pearson (r) correlation coefficient between \textit{LLM-predicted ratings} and the \textit{averaged human ratings} across items as the primary evaluation metric, and root mean squared error (RMSE) as the auxiliary metric. Higher Pearson correlation and lower RMSE suggest higher alignment with human interpretations of gender associations.

For a clearer interpretation of LLMs' correlation performance in consideration of inherent subjectivity in the human ground truth ratings, we also adopted two human performance baselines by using the uncertainty measures from \autoref{sec:human_uncertainty}: (i) the leave-one-rater-out (LOO) correlation corresponds to the agreement level of a typical human rater, while (ii) the intraclass correlation coefficient $\sqrt{\mathrm{ICC}(1,k)}$ approximates the maximal attainable correlation with the aggregated human judgments (i.e., an ideal model that perfectly recovers the latent true rating from the human population, so that the error only comes from measurement noise). The two human baselines suggest the correlation ceilings for the model evaluation due to rater uncertainty and measurement noise.

\vspace{-0.5em}
\subsection{Results}
\vspace{-0.5em}


\paragraph{Overview of Model Alignment}

\begin{figure}[htbp]
    \centering
    \vspace{-5pt}
    \includegraphics[width=\textwidth]{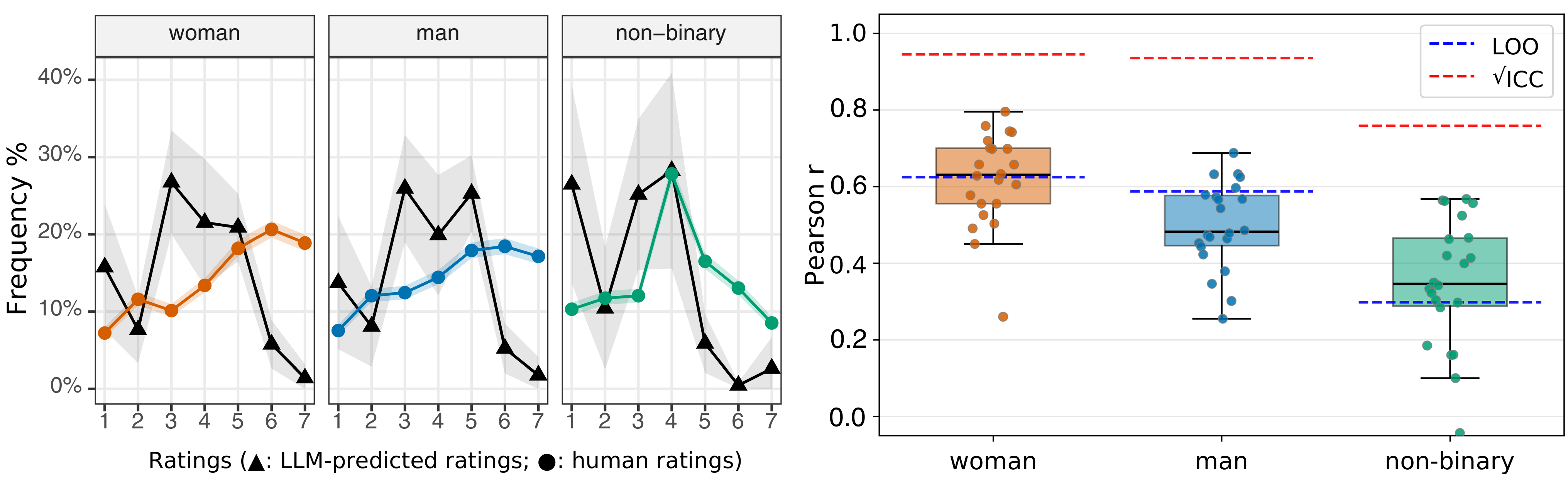}
    \caption{\textit{Left}: the frequency\% distribution of predicted rating scores (1-7) per gender category of all LLMs, compared to human ratings. \textit{Right}: Pearson r (correlation between LLM and averaged human ratings) distribution of all models per gender category; each point represents an evaluated LLM.
    }
    \vspace{-10pt}
    \label{fig:model_collective}
\end{figure}

We first examine how well state-of-the-art (SOTA) language models align with human judgments about how gender is associated with physical attributes in \dataset\ at the collective level. \autoref{fig:model_collective} (right) presents model–human correlations alongside the two reliability-based reference baselines. Overall, SOTA language models show moderate alignment with human gender associations across attributes. However, alignment varies systematically across gender categories. In particular, models tend to achieve higher correlations for \textit{woman and non-binary person} than for \textit{man}, centering around or exceeding the LOO (typical-rater) baseline. Interpreting these results relative to the reliability ceilings helps account for measurement noise, suggesting that the observed differences reflect genuine gender variation in model–human alignment in SOTA LLMs: SOTA LLMs are better aligned with humans in interpreting the semantic association of physical attributes with woman and non-binary person than the male category. 

Notably, model rating distributions are more concentrated in the mid-range (3–5) and less frequent at the extremes (6–7) compared to human ratings, which are more evenly distributed across the scale (\autoref{fig:model_collective}, left). This pattern suggests that models tend to avoid extreme judgments, producing more conservative or less differentiated ratings than human annotators, and thus only partially capture the strength of gender associations present in human interpretations.


 \begin{figure}[htbp]
\centering
\begin{subfigure}[t]{0.49\linewidth}
    \centering
    \includegraphics[width=\linewidth]{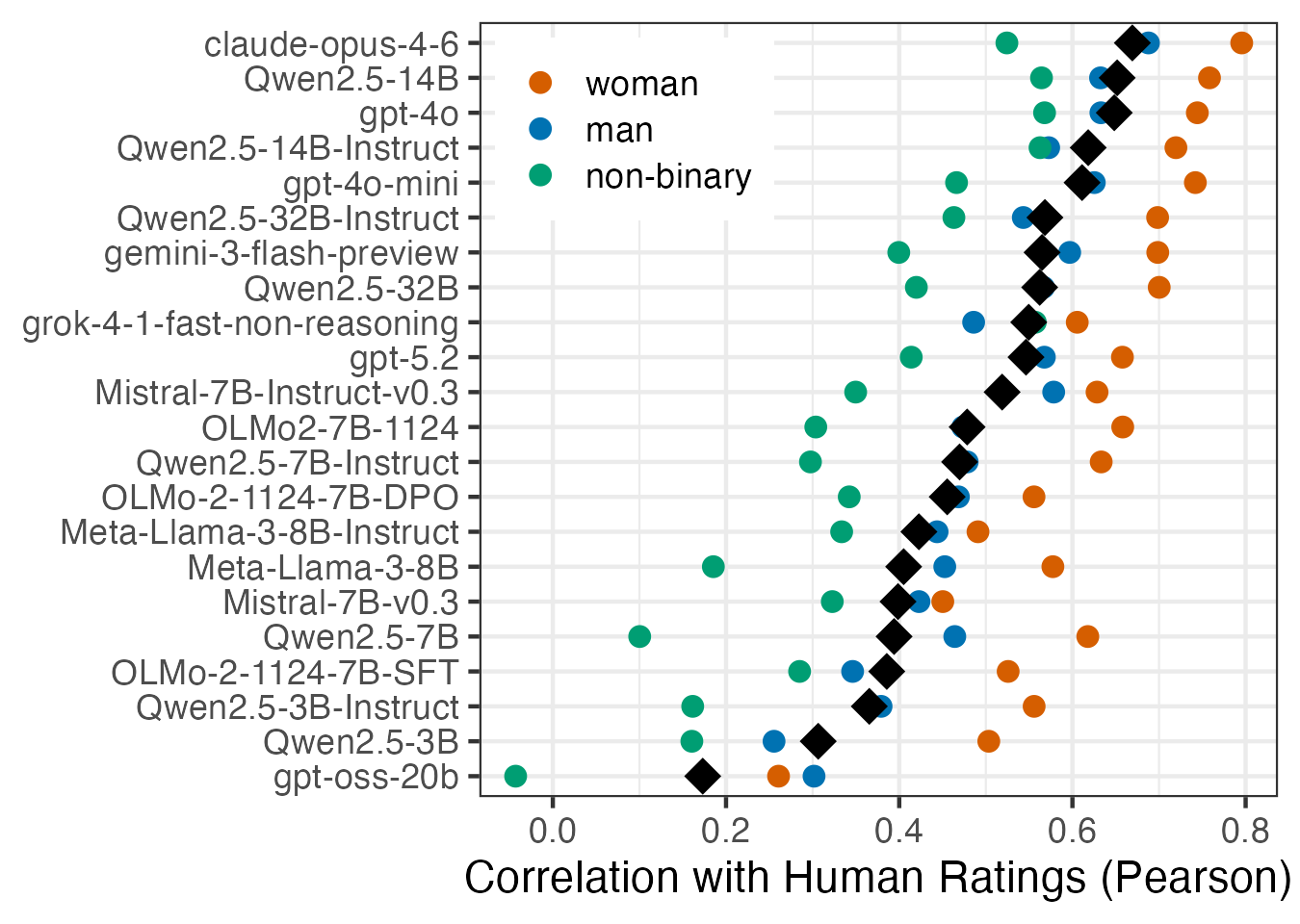}
    \caption{Correlation results of all evaluated models on \dataset. \filleddiamond~means the overall average.}
    \label{fig:model_r_all}
\end{subfigure}
\hfill
\begin{subfigure}[t]{0.49\linewidth}
    \centering
    \includegraphics[width=\linewidth]{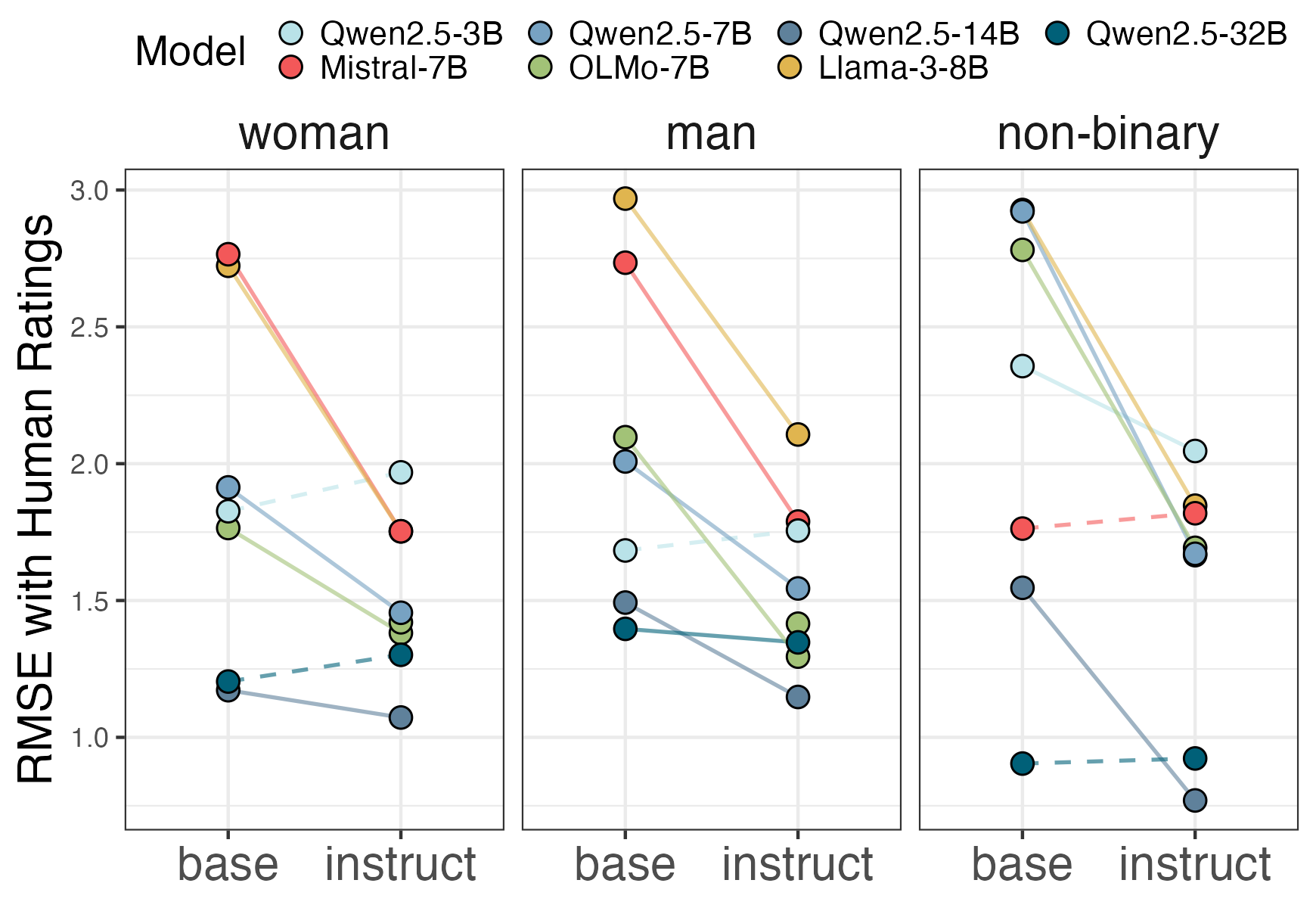}
    \caption{The comparison of model-human alignment (RMSE) between base and instruct models.}
    \label{fig:model_base-instruct}
\end{subfigure}

\vspace{6pt}

\begin{subfigure}[t]{\linewidth}
    \centering
    \includegraphics[width=\linewidth]{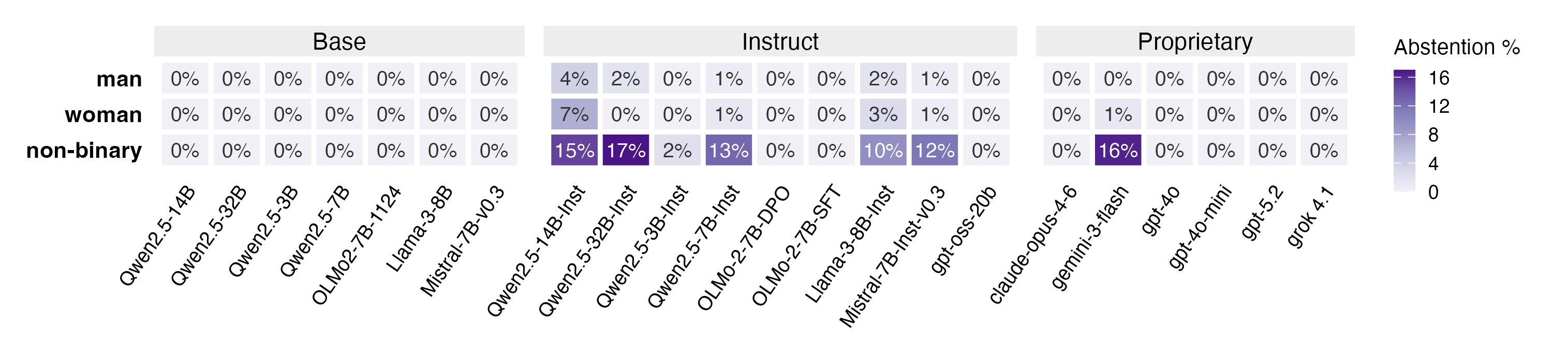}
    \caption{Model heatmap of abstention responses on \dataset. Instruct models show significant abstention rates among models; abstention is significantly more likely to occur for the non-binary gender category.}
    \label{fig:model_abstention}
\end{subfigure}
\caption{Zero-shot inference experiment results of the state-of-the-art LLMs.}
\vspace{-1em}
\end{figure}

\vspace{-0.5em}
\paragraph{Model Comparisons} \autoref{fig:model_r_all} demonstrates the results of model-human alignment in gender associations of physical attributes evaluated on the complete set of 16 models. Consistent with \autoref{fig:model_collective} (right), gender differences remain systematically salient within each model's performance. Two other results stand out. First, Claude-Opus-4.6 achieves the strongest alignment with human gender associations, with an average Pearson correlation of r = 0.669. In contrast, GPT-OSS-20B has surprisingly the weakest overall alignment (r = 0.173) and even shows a negative correlation for the non-binary category (r = -0.043). Second, compared with prior gender-bias benchmarking work \citep{yang2025demographics,liang2022holistic, felkner2023winoqueer}, proprietary models do \textit{NOT} show a clearly dominant separation from open-source models on \dataset. At a collective level, proprietary models show only a modest advantage, with substantial overlap in practical performance rather than a distinct lead.


We also directly compare base and instruct variants of the same model backbone, as illustrated in \autoref{fig:model_base-instruct}. Instruction-tuned variants of open-source models generally improve over their base counterparts in RMSE. This pattern suggests that post-training alignment reduces absolute prediction divergence from the human ground-truth. However, no comparable gains are consistently observed in Pearson~$r$, indicating that it does not substantially improve the rank-order alignment of attribute–gender associations.

\vspace{-1em}
\paragraph{Abstention Analysis}
We also investigate to what extent LLMs abstain from answering the question. We count a model response to be an abstention if it specifically expresses an abstention, such as  ``physical attributes are not related to gender identity'', and ``non-binary people can have any eye shape, just like anyone else.'' In such abstention cases, models either choose the lowest 1 (not likely at all) as the rating to express refusal and disagreement on the question assumption, or simply end the response without a rating.

We conduct a more systematic quantitative analysis on abstention counts by analyzing the model responses, which reveal two salient patterns (\autoref{fig:model_abstention}). Firstly, we observed highly asymmetric abstention in the model responses against the non-binary gender category.
This indicates that the burden of refusal or deflection is not uniform across gender groups, but is disproportionately concentrated on the non-binary category. This finding crucially warns of practical accessibility bias caused by over-triggering of sensitivity and safety flags, and the behavior of AI refusal disproportionately constraining more vulnerable groups \citep{luo2024refusal, abramovich2024refusal}.
Secondly, abstention is not a broadly shared behavior across all systems. Rather, it is concentrated almost entirely in the proprietary model Gemini-3-flash and instruct-tuned open-source models, while the corresponding base models show little to no comparable tendency. This distribution suggests that abstention is not simply a property of model scale or architecture, but is closely associated with post-training and alignment procedures. Taken together, these patterns are consistent with the interpretation that post-training can introduce asymmetric over-sensitivity, causing some models to over-trigger safety or sensitivity heuristics specifically in response to non-binary-targeted prompts.



\section{Scalable Socio-Linguistic Analyses Using a Trained Proxy Predictor}
\label{sec:scaleup}

\vspace{-0.5em}
\subsection{Training a Proxy Model for Predicting Humans' Gender Associations}
\vspace{-0.5em}

To extend the contribution of this study beyond human experiments and model evaluation analyses, we further developed a proxy predictor model on the full \dataset's data by selecting and training from the open-source models evaluated in \autoref{sec:model_exp}.

\begin{wrapfigure}{r}{0.48\textwidth}
    \centering
    \vspace{-10pt}
    \includegraphics[width=0.46\textwidth]{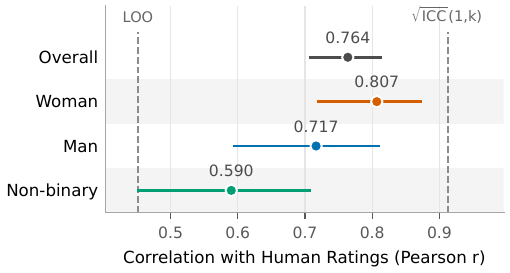}
    \caption{Training results of the predictor model, evaluated on GAPA (test).}
    \label{fig:novels-plot}
    \vspace{-10pt}
\end{wrapfigure}

To seek the best model candidate for training the proxy predictor, we first conduct supervised fine-tuning across an extensive search grid of LLMs and hyperparameters on \dataset~(60/15/25 split for train/validation/test). To match the prediction goal, each model is augmented with a linear regression head to directly map aggregated human ratings for gender categories $c \in \{\textit{woman}, \textit{man}, \textit{non-binary person}\}$. Given an input sentence describing an attribute for a target gender (e.g., ``a woman has broad shoulders''), the mean-pooled final-layer representation is projected to a scalar score, framing the task as continuous regression to directly assess model–human alignment. The training optimizes mean squared error (MSE) on averaged human ratings, encouraging recovery of shared annotator signals while reducing individual noise.

For a fair comparison, we use a standardized hyperparameter search with model-specific ranges (e.g., learning rate, batch size), and select the best-performing configuration for each model. All models share the same tuning budget and are evaluated under consistent cross-validation and multi-seed settings. Full details including design choices and hyperparameter specifications are provided in \autoref{app:model_training}.

The final proxy model takes the backbone of \texttt{olmo2\_7b\_base}, and achieves the highest test-set correlation of $r=0.764$ and lowest RMSE $=0.633$. This predictor can serve as a simulator for a scalable collection of human-like gender-association interpretations on unseen items, supporting larger-scale sociolinguistic analysis.

\subsection{Scaled-Up Analyses on LitBank \citep{litbank_bamman2019annotated}}
\label{subsec:novels}

\begin{wrapfigure}{r}{0.46\textwidth}
    \centering
    \includegraphics[width=0.44\textwidth]{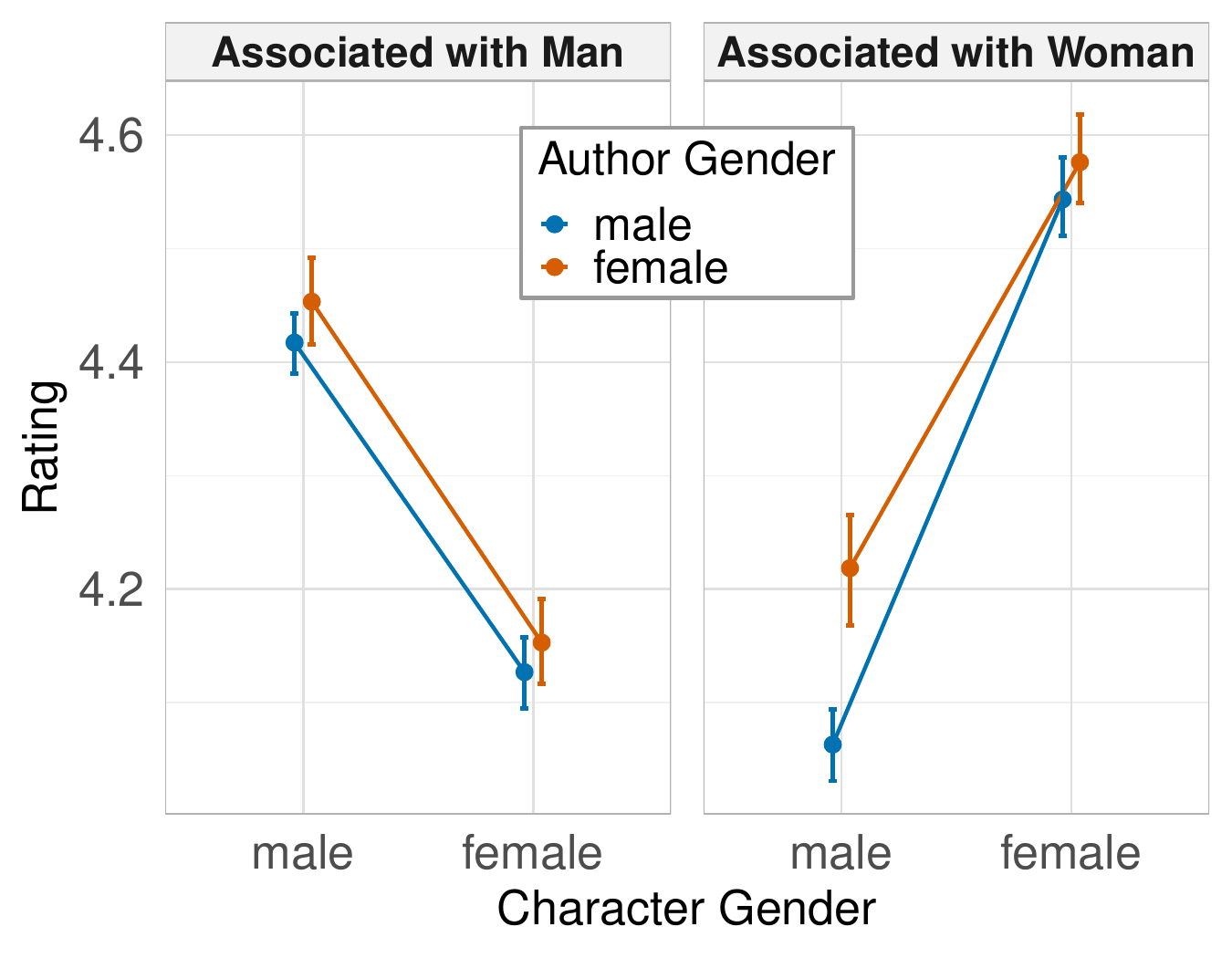}
    \caption{Predicted gender association ratings of physical attributes describing LitBank
    characters, by character gender and author gender.}
    \label{fig:novels-plot}
\end{wrapfigure}

As a preliminary demonstration of what the predictor model can do at scale, we utilize the extraction pipeline previously used in  \autoref{sec:dataset} to extract physical attributes from LitBank \citep{litbank_bamman2019annotated, sims2019literary_litbank}, a literature corpus of 100 novels from Project Gutenberg, which consists of fiction (both novels and short stories) in a mix of genres and literary styles before 1923. After extracting the attributes, we employed the proxy model to estimate the gender association ratings of each attribute with men, women, and non-binary people, yielding a three-dimensional gender-association profile for every attribute. This enabled us to conduct exploratory analyses of their relationships with two types of metadata: author gender and the gender of the character each physical attribute describes in the original novel. We primarily explore the hypothesis\footnote{The analyses reflect findings of contemporary reader perceptions in the US due to the participant demographics of \dataset. See \autoref{app:novels} for a more detailed interpretation.}: the physical descriptions used to describe a literary character are systematically gendered on their own; their gender associations can be predicted by who the writer is (author gender), and about whom the description is written (character gender).


\paragraph{Character Gender.}
As shown in \autoref{fig:novels-plot}, physical descriptions differentiate male and female
characters through the direction rather than the magnitude of their gender associations.
Attributes describing male characters are rated as more associated with man than with
woman, and those describing female characters show the reverse ordering, but the two groups
are comparably distinctive in absolute terms.
vs.\ $0.71$, $p = .20$).

\paragraph{Author Gender.}
Author gender modulates this structure along only one of the two binary dimensions.
Ratings of association with man are largely invariant to the author gender,
and female characters elicit similar
descriptions from both groups of authors. Male characters do not: male authors describe men
using attributes with markedly weaker feminine associations than female authors do ($4.06$
vs.\ $4.22$; $\beta = -0.12$, $p = .03$). Authorial variation is thus confined to the degree
of feminine association admitted into the description of a man, while masculine association
follows a convention shared across author groups. The results carry the central claim of this work from controlled annotation into naturalistic text at scale:
descriptions that seem independent of gender assignment remain systematically gendered by whom they describe and by who describes them.

\section{Conclusion}
\vspace{-0.4em}

This study challenges a widely held assumption in AI fairness, accessibility, and ethics research: that \textit{replacing categorical identity labels with ``objective'' physical descriptions yields gender-neutral communication}. Introducing GAPA, a large-scale human dataset of gender associations for 316 physical attributes, we show that \textbf{this assumption does not hold}: physical descriptions carry automatic, structured, and graded gender associations among human readers, with over half of attributes significantly more associated with one gender than others. Evaluating 16 LLMs against these human ratings, we find that models partially recover human patterns yet with systematic misalignment: model ratings are compressed toward the midpoint, they align less well with human judgments of the man category, and instruction-tuned and proprietary models show asymmetric abstention that disproportionately targets the non-binary category, raising distinct equity concerns. Finally, we developed a proxy predictor to analyze gendered descriptions beyond controlled annotation settings, but at scale in naturalistic text. Together, these findings provide the first empirical evidence that ``objective'' physical descriptions do not neutralize gender, but instead shift gendered communication into a less visible register with downstream consequences for how models should be evaluated and designed for human–AI interaction.
\vspace{-0.5em}

\section*{Limitations}
\vspace{-0.4em}
Several limitations qualify the human norming results. First, the task measures socially shared associations between physical attributes and gender categories, rather than whether attributes are inherently gendered. Ratings may therefore reflect stereotype knowledge, perceived prevalence, and expectations about language use in addition to semantic association.

Second, the attribute set is heterogeneous, spanning bodily morphology, grooming, accessories, age related cues, and figurative descriptions that may signal gender through different mechanisms. Attributes were also evaluated in isolation, without the discourse context that can shape gendered interpretation. Finally, \textit{non-binary person} may have less culturally shared appearance-based prototypes than \textit{woman} or \textit{man}, so lower agreement may reflect greater conceptual heterogeneity. The findings should therefore be interpreted as evidence of structured \emph{socially shared gender associations in physical description}, rather than a complete account of gendered meaning in language.


\section*{Acknowledgments}


This work was supported in part by the UCLA Graduate Summer Research Mentorship (GSRM) Program.
We also thank the members of the UCLA Coalas Lab for their valuable comments throughout this project. Special thanks go to Hillary Nguyen, Ella Han, Troy Tian, Jonathan Pak, and Ryan Zheng for their assistance with manually verifying the novel-extracted attributes in \dataset~to ensure the quality of automated attribute extraction from novels. We further thank Rohan Jain, Maureen Widjaja and Karin Yamaoka for their feedback early on in the project. Finally, we're grateful to Google's GiG and GCP Credit Program for their support of this research.

\section*{Ethics Statement}

This study examines socially shared gender associations in linguistic descriptions of physical attributes. We emphasize that our work does not define or infer individuals’ gender from appearance, but instead analyzes patterns in language and perception. Because such analyses risk reinforcing stereotypes if misused, we frame all findings descriptively and caution against using the dataset or models for gender inference, profiling, or decision-making about real individuals.

Human-subject data were collected with informed consent, fair compensation, and no personally identifiable information beyond basic demographics. All analyses are conducted at the aggregate level. The dataset and models are released for research purposes, such as auditing implicit bias and improving AI transparency, and should be used with appropriate safeguards and contextualization.
\bibliography{colm2026_conference}
\bibliographystyle{colm2026_conference}

\newpage

\appendix

\section{Supplementary Information of Dataset Collection for \dataset}
\label{app:attrselection}

Physical attributes were collected from three distinct source domain to facilitate both large coverage, cognitively prominent concepts, and rare and stylized descriptors.

\paragraph{LLM-generated attributes} were collected by prompting GPT-4o and Gemini 2.0 to list physical attributes that follow the pattern of ``\{ADJ\} + \{NOUN\}'', where \{ADJ\} describes the properties of a body part that is referred to by \{NOUN\} (e.g., ``an hourglass figure'', ``a chiseled jawline'').

\paragraph{Human-written attributes} were collected by presenting participants with a diverse set of images depicting individuals whose facial appearance contained both stereotypically gender-typical and gender-atypical features: \textit{N} = 26 images generated by GPT 4o and \textit{N} = 22 naturalistic images selected from the Flickr-Faces-HQ dataset \citep{karras2019flickr-faces}. During the study, each participant saw a total of 10 images. Participants viewed one image at a time and were asked, ``How strongly do you associate this person’s appearance with that of a woman/man/non-binary person in your culture?'' using a 7-point scale. After providing their ratings, participants were prompted to list three features of the person’s appearance that led them to associate the individual with the gender category they rated highest. They were also given the option to identify at least one feature of the person’s appearance that seemed unusual or atypical for someone they would associate with the gender category they rated highest.

\paragraph{Attributes extracted from novels} were collected using a specially designed extraction pipeline for sourcing physical attributes from any long-context documents or literature corpora such as novels. The design and implementation details of the extraction pipeline is elaborated in \autoref{app:extraction}. To construct the novel-extracted set of the current dataset, we chose six popular contemporary novels of similar size, genre, and a balanced woman-man author ratio: \textit{Game of Thrones} (Vol.\ 1), \textit{Harry Potter} (Vol.\ 1), \textit{Twilight} (Vol.\ 1), \textit{Hunger Games} (Vol.\ 1), \textit{Lord of the Rings} (Vol.\ 1), and \textit{Maze Runner}. From each book source, we randomly sampled 16 attributes from the extracted physical attributes. 




\section{Supplementary Information for Human Annotations on \dataset}
\label{app:humman_info}

\subsection{Participant Demographics}
\label{app:participant-demographics}

The study was certified as exempt from IRB review by the UCLA Office of the Human Research Protection Program (IRB-24-5529; 45 CFR 46.104). All participants provided informed consent prior to participation. We recruited 320 US participants via Prolific (\$12/hour). Recruitment criteria required participants to be native English speakers, residing in the United States, with a Prolific approval rate of at least 95\% and passing an attention check.
The final sample included 75 female. The participants ranged in age from 19 to 83 years (M = 42.70).

\subsection{Descriptive Results}

\autoref{tab:human_descriptives} reports descriptive statistics of human-rated gender likelihood for attributes across three sources (human-written, LLM-generated, and novel-derived) for each target gender. A consistent pattern emerges across sources: ratings are highest for women, followed by men, and lowest for non-binary targets. This ordering holds within most individual sources and in the overall aggregates. This suggests a systematic difference in how our collected attributes are overall associated with gender categories in human judgments.

\begin{table}[htbp]
\centering
\footnotesize
\caption{Descriptive statistics by gender and attribute source}
\label{tab:human_descriptives}
\setlength{\tabcolsep}{3.5pt}
\begin{tabular}{llrrrrr}
\toprule
\textbf{Gender} & \textbf{Source} & $n$ & Mean & SD & CI Low & CI High \\
\midrule
\multirow{4}{*}{Man}
  & Human & 744  & 4.27 & 1.83 & 4.14 & 4.40 \\
  & LLM   & 2760 & 4.25 & 1.85 & 4.19 & 4.32 \\
  & Novel & 1364 & 4.38 & 1.93 & 4.28 & 4.48 \\
\cmidrule(l){2-7}
  & \textit{\textbf{Overall}} & 4868 & 4.29 & 1.87 & 4.24 & 4.34 \\
\midrule
\multirow{4}{*}{non-binary}
  & Human & 762  & 4.05 & 1.55 & 3.94 & 4.16 \\
  & LLM   & 2766 & 4.07 & 1.66 & 4.01 & 4.14 \\
  & Novel & 1354 & 3.76 & 1.85 & 3.66 & 3.86 \\
\cmidrule(l){2-7}
  & \textit{\textbf{Overall}} & 4882 & 3.98 & 1.71 & 3.93 & 4.03 \\
\midrule
\multirow{4}{*}{Woman}
  & Human & 794  & 4.20 & 1.93 & 4.07 & 4.33 \\
  & LLM   & 2880 & 4.71 & 1.80 & 4.64 & 4.77 \\
  & Novel & 1282 & 4.24 & 1.97 & 4.13 & 4.35 \\
\cmidrule(l){2-7}
  & \textit{\textbf{Overall}} & 4956 & 4.50 & 1.88 & 4.45 & 4.56 \\
\bottomrule
\end{tabular}
\end{table}

\autoref{fig:attribute_results} illustrates the per-gender ratings of all attributes with significant gender differences ($p < 0.05$) in the complete six ranking patterns.

\begin{figure}[tbp]
    \centering

    \begin{subfigure}[t]{0.48\textwidth}
        \centering
        \includegraphics[
            width=\linewidth,
            height=0.78\textheight,
            keepaspectratio
        ]{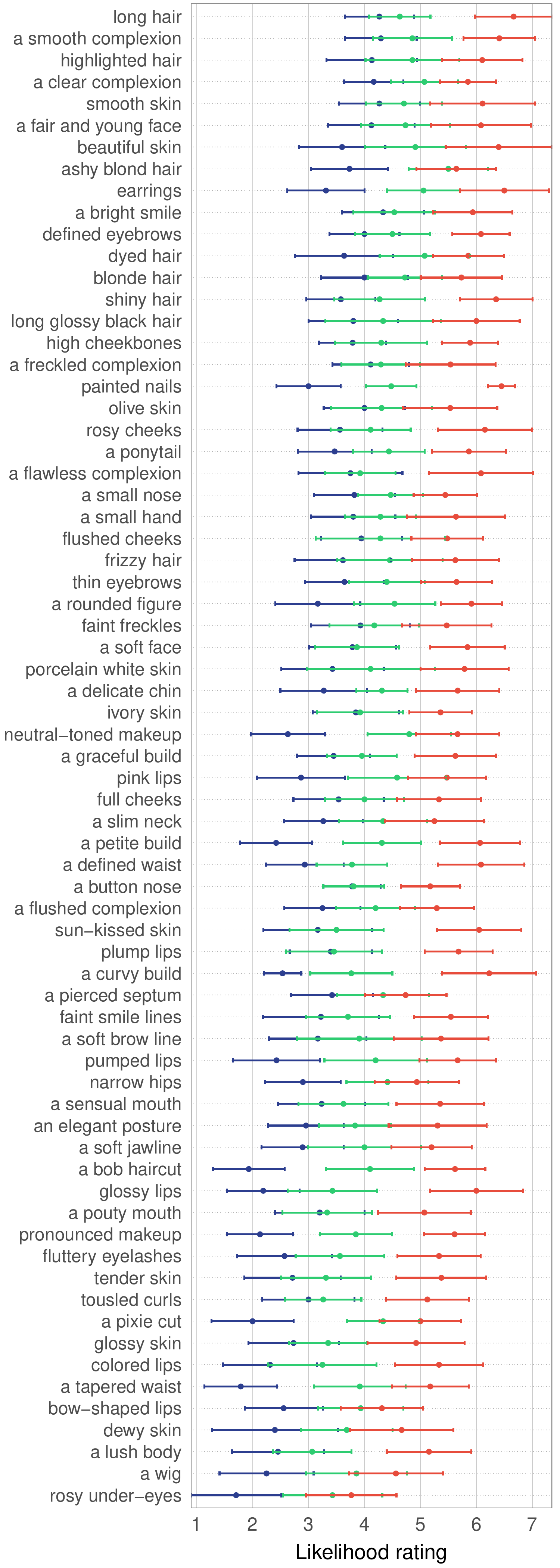}
        \caption{woman $>$ non-binary person $>$ man}
        \label{fig:attr_w_n_m}
    \end{subfigure}
    \hfill
    \begin{subfigure}[t]{0.48\textwidth}
        \centering
        \includegraphics[
            width=\linewidth,
            height=0.78\textheight,
            keepaspectratio
        ]{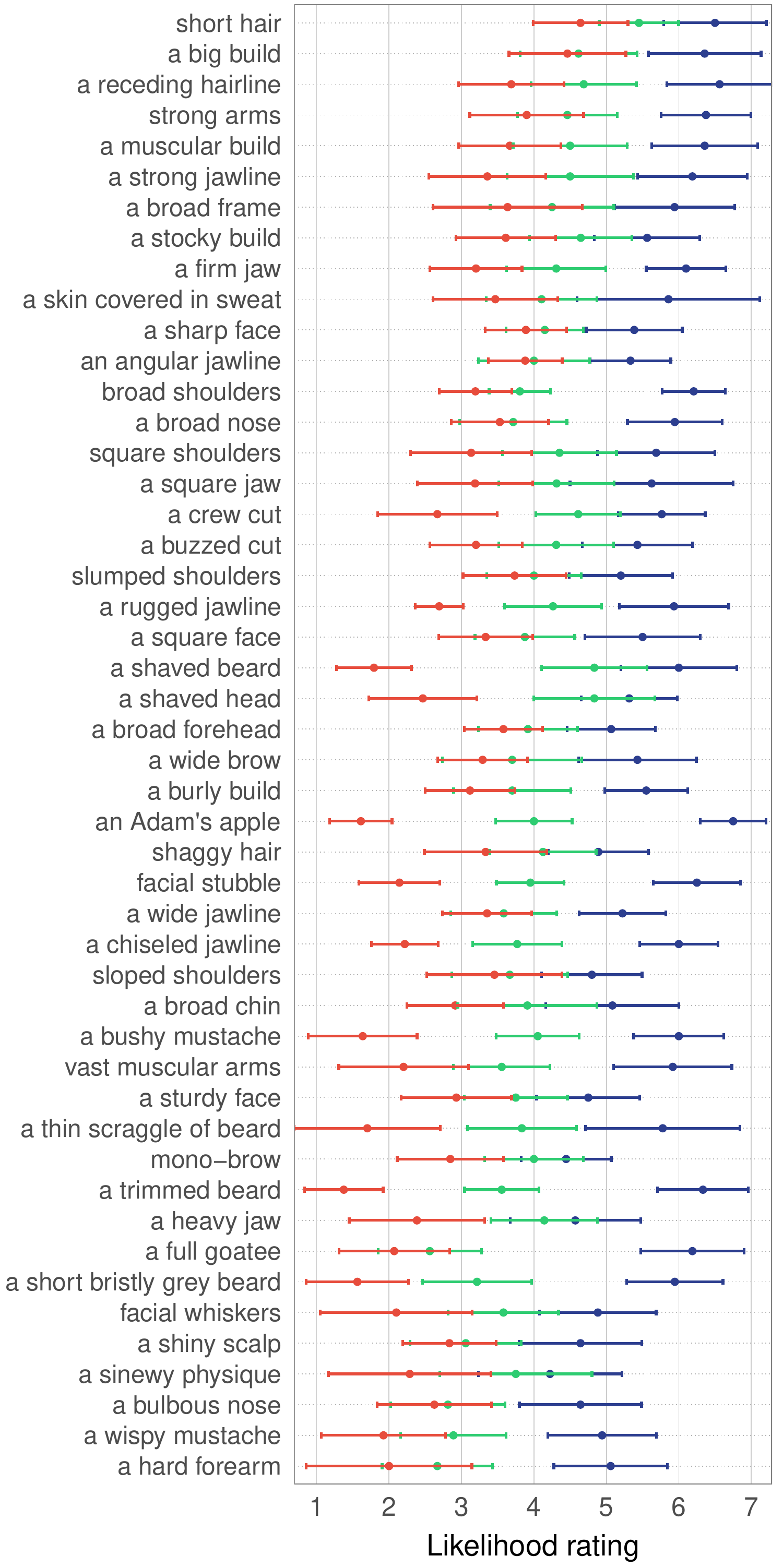}
        \caption{man $>$ non-binary person $>$ woman}
        \label{fig:attr_m_n_w}
    \end{subfigure}

    \caption{Per-gender mean ratings and 95\% confidence intervals for
    attributes showing significant gender differences ($p < 0.05$), grouped
    by ranking pattern.}
    \label{fig:attribute_results}
\end{figure}

\begin{figure}[tbp]
    \ContinuedFloat
    \centering

    \begin{subfigure}[t]{0.48\textwidth}
        \centering
        \includegraphics[
            width=\linewidth,
            height=0.78\textheight,
            keepaspectratio
        ]{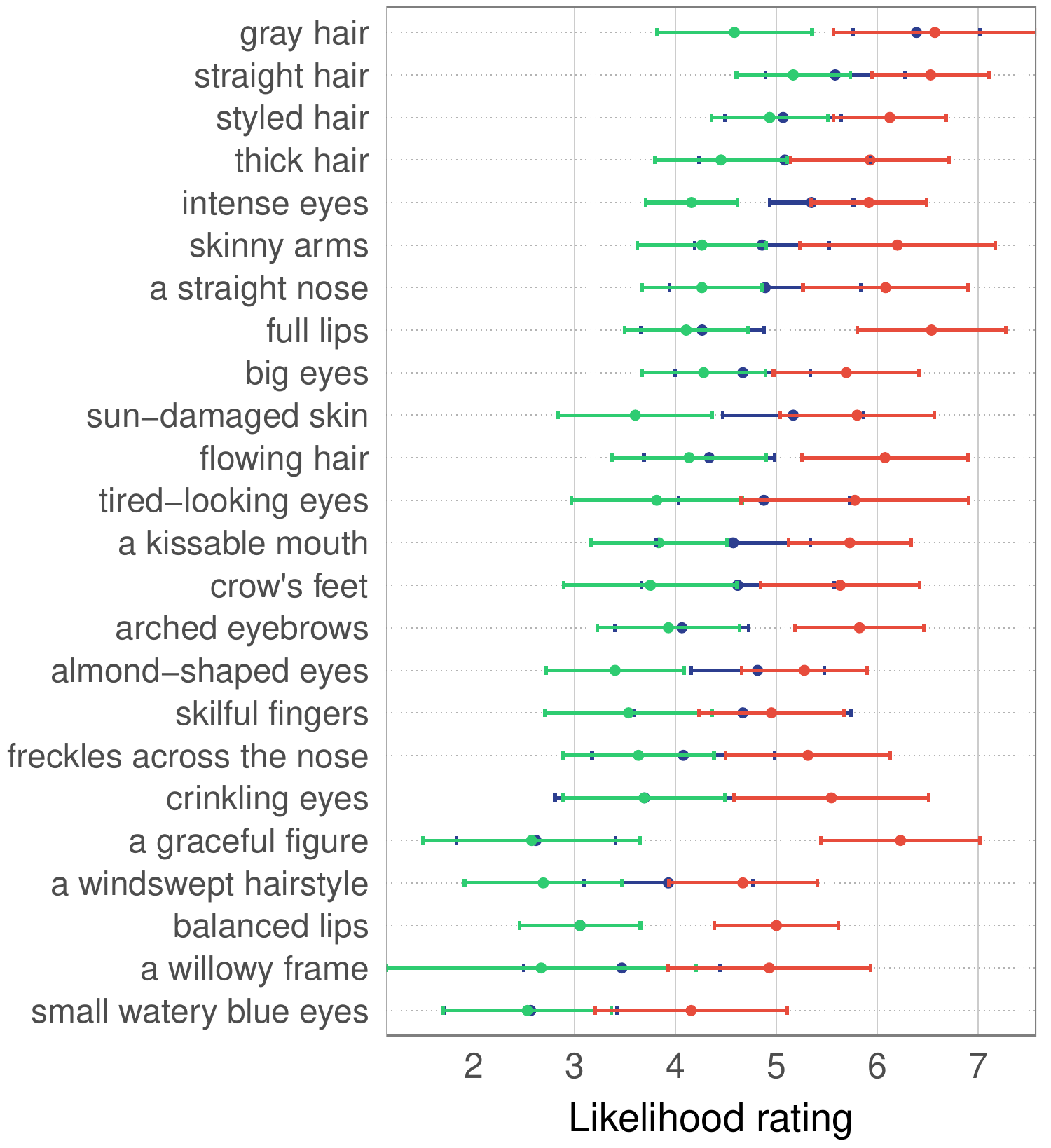}
        \caption{woman $>$ man $>$ non-binary person}
        \label{fig:attr_w_m_n}
    \end{subfigure}
    \hfill
    \begin{subfigure}[t]{0.48\textwidth}
        \centering
        \includegraphics[
            width=\linewidth,
            height=0.78\textheight,
            keepaspectratio
        ]{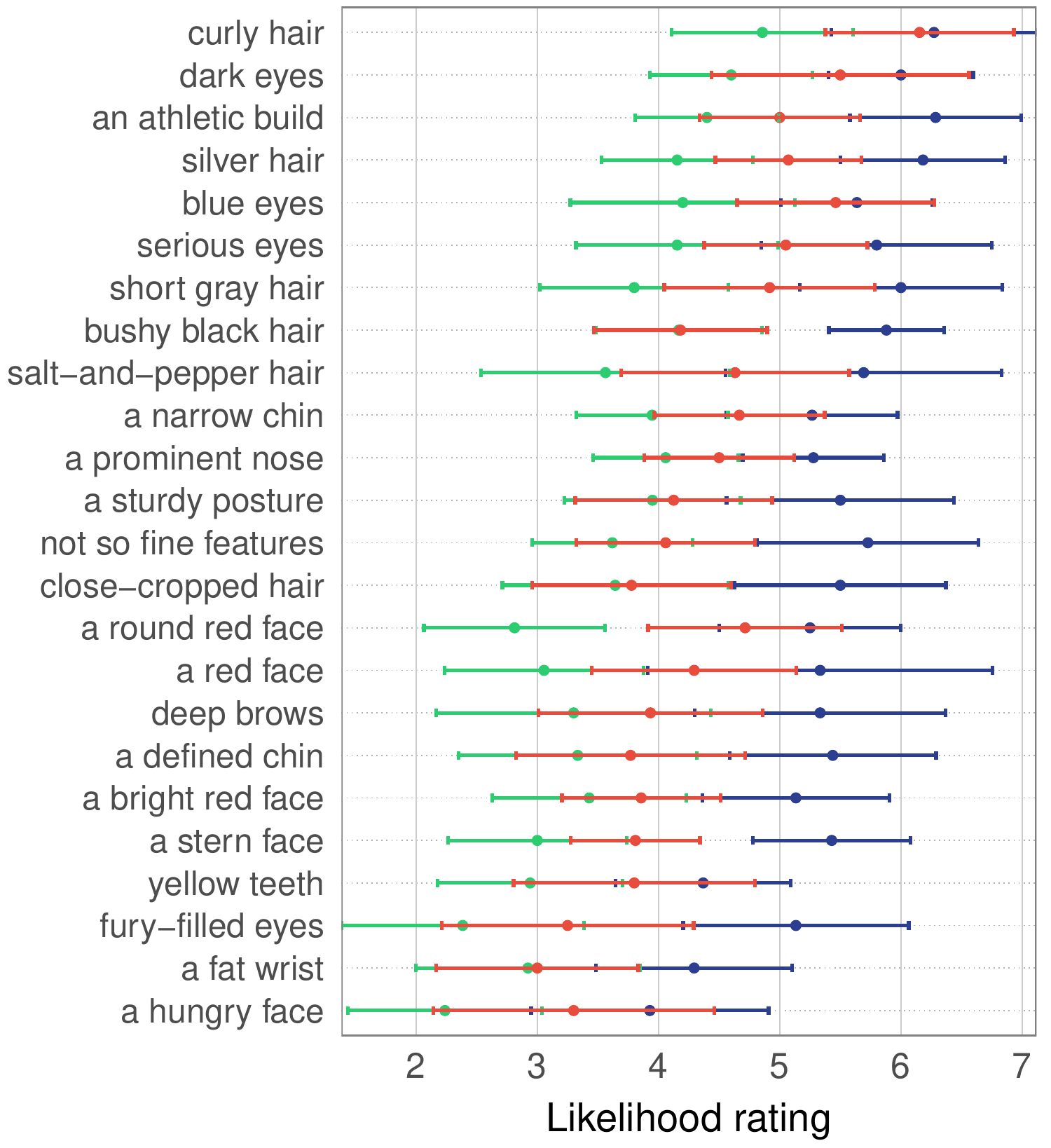}
        \caption{man $>$ woman $>$ non-binary person}
        \label{fig:attr_m_w_n}
    \end{subfigure}

    \caption{Per-gender ratings grouped by ranking pattern (continued).}
\end{figure}

\begin{figure}[tbp]
    \ContinuedFloat
    \centering

    \begin{subfigure}[t]{0.48\textwidth}
        \centering
        \includegraphics[
            width=\linewidth,
            height=0.78\textheight,
            keepaspectratio
        ]{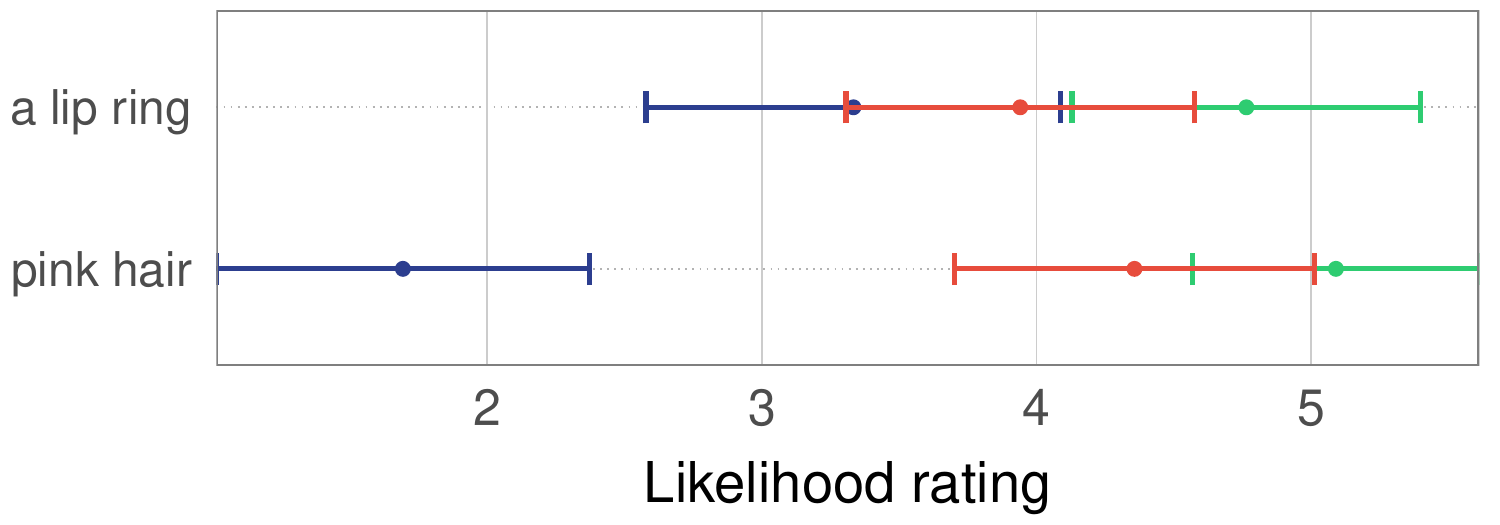}
        \caption{non-binary person $>$ woman $>$ man}
        \label{fig:attr_n_w_m}
    \end{subfigure}
    \hfill
    \begin{subfigure}[t]{0.48\textwidth}
        \centering
        \includegraphics[
            width=\linewidth,
            height=0.78\textheight,
            keepaspectratio
        ]{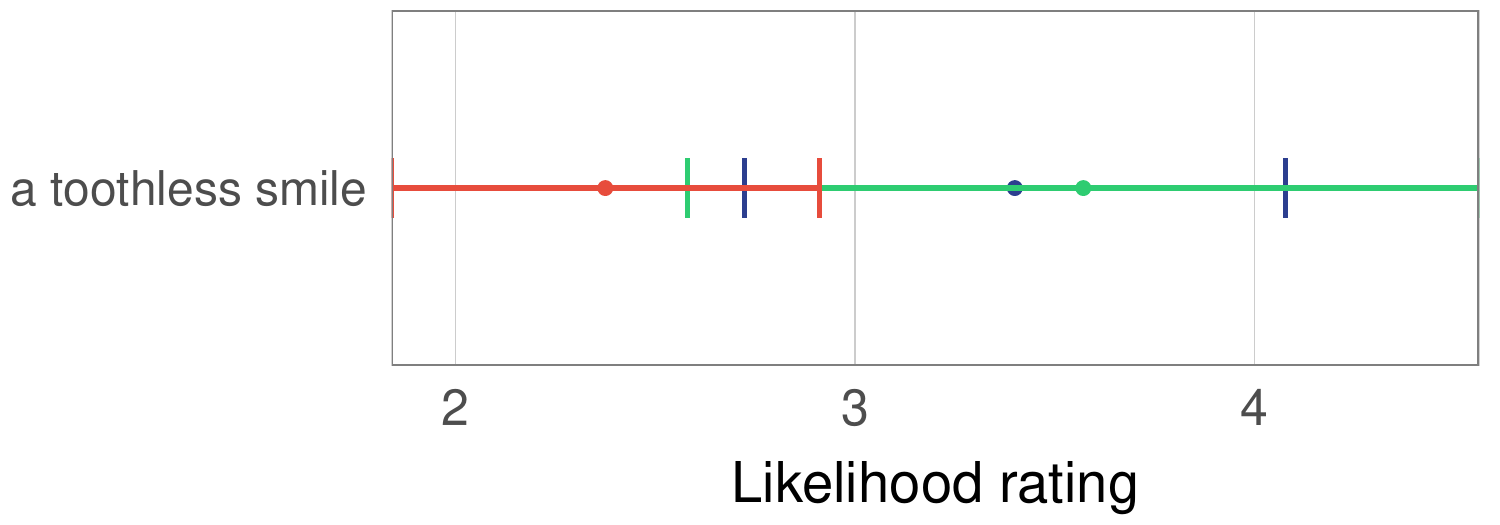}
        \caption{non-binary person $>$ man $>$ woman}
        \label{fig:attr_n_m_w}
    \end{subfigure}

    \caption{Per-gender ratings grouped by ranking pattern (continued).}
\end{figure}

\subsection{Correlation Ceilings from Human Uncertainty}
\label{app:noise-ceiling}
As the target labels $\bar{y}_i$ are derived from the average of noisy human judgments that are highly subjective and diverse (see~\autoref{sec:human_exp_results}), \textbf{correlation performance is fundamentally bounded by inter-rater variability} \citep{cronbach1972dependability}. We therefore estimate two complementary noise ceilings: a \emph{leave-one-rater-out} (LOO) ceiling and an \emph{intraclass correlation} (ICC) ceiling, illustrated in \autoref{fig:noise-ceiling}. They later serve as interpreting references for models of different strengths \autoref{fig:model_collective} (right).

\paragraph{Leave-One-Out Reliability Ceiling.} The LOO ceiling estimates the agreement between a typical individual rater and the consensus of other raters. For each rater $j$, we compute $r_j^{\text{LOO}}$, the Pearson correlation between that rater’s ratings and the mean ratings of all remaining raters across items:
\[
r_j^{\text{LOO}} = \mathrm{corr}\big(\{y_{ij}\}, \{\bar{y}_{i,(-j)}\}\big),
\]
\[
\text{where \quad} \bar{y}_{i,(-j)} = \frac{1}{k-1}\sum_{j' \neq j} y_{ij'}.
\]
The LOO ceiling $r^{\text{LOO}}$ is then obtained by averaging $r_j^{\text{LOO}}$ across raters. This ceiling value reflects \emph{human-level performance}: a model (with parameter $\theta_{typical}$) achieving LOO-level correlation performs at the same level as any \textit{typical} individual annotator. Consequently, a model with a comparable performance to a typical human rater with idiosyncratic rater noise is expected to satisfy
\[
r(\theta_{typical}) \; \sim \; r^{\text{LOO}}.
\]

\paragraph{Intraclass Correlation (ICC) Ceiling.}
In contrast, the intraclass correlation coefficient ICC$(1,k)$ estimates the reliability of the \emph{mean} rating across $k$ raters, describing the distribution of the collected human rating data. ICC is a variance-partitioning reliability coefficient derived from a random-effects model and is widely used to quantify inter-rater reliability in behavioral measurement \citep{shrout1979intraclass, mcgraw1996forming}. Under a standard random-effects model,
\[
y_{ij} = \tau_i + \varepsilon_{ij},
\]
where $\tau_i$ denotes the \textit{latent true attribute} value for item $i$ and $\varepsilon_{ij}$ represents rater-specific noise assumed to be independent with mean zero, the population-level ICC for the mean of $k$ raters is:
\[
\mathrm{ICC}(1,k) =
\frac{\sigma_\tau^2}{\sigma_\tau^2 + \sigma_\varepsilon^2 / k}.
\]
ICC$(1,k)$ represents the proportion of variance in the averaged human ratings attributable to the latent signal rather than noise, which has been adopted as the reliability of the averaged human judgment across psychometrics literature \citep{shrout1979intraclass, cronbach1972dependability}.

Based on classical measurement theory, correlation between a noisy measurement and any predictor is bounded by the square root of that measurement’s reliability (i.e., known as attenuation due to measurement error) \citep{spearman1961proof, mcgraw1996forming}. Consequently,
ICC$(1,k)$ provides a \emph{natural upper bound} on the attainable correlation of an ideal model (with parameter $\theta_{ideal}$) that does not contain any rater noise in any rating prediction with the averaged human ratings. In other words, ICC$(1,k)$ describes the correlation of an ideal model that perfectly recovers the latent true rating from the human population (by always predicting $\tau_i$ i.e., $\hat{y}_i(\theta_{ideal}) = \tau_i$)), offering a stricter theoretical noise ceiling than $r^\text{LOO}$ for correlation-based evaluation:

\[
r(\theta_{ideal}) \sim \mathrm{corr}(\tau_i, \bar{y}_i) = \sqrt{\mathrm{ICC}(1,k)}.
\]

While $r^\text{LOO}$ and $\sqrt{\mathrm{ICC}(1,k)}$ provide the theoretical ceiling references for interpreting model correlation values, it is important to note that the empirical results of model performances may occasionally exceed them, due to the sampling noise in the finite set of collected data for evaluation.

\section{Extraction Pipeline}
\label{app:extraction}

We describe the data source, the extraction pipeline implemented, and the schema and features of the extracted attributes used in our experiments.



\subsection{Extraction Pipeline Process}
\label{app:extraction-process}
\paragraph{1. Lexicon-based sentence filter.}
Sentences are obtained with NLTK's \texttt{sent\_tokenize}. We then restrict LLM calls to sentences that are likely to contain physical descriptions. Two hand-defined lexicons are used: (a)~\textbf{physical-attribute nouns}: body-part terms (e.g., head, face, hair, eyes, nose, jaw, hands, skin, build, posture) and related nouns (e.g., silhouette, frame); (b)~\textbf{physical-attribute adjectives}: descriptors for build/size (e.g., thin, broad, tall), musculature (e.g., muscular, toned), face and eyes (e.g., sharp, round, deep-set), hair (e.g., curly, blond), skin (e.g., freckled, pale-skinned), and evaluative terms (e.g., beautiful, handsome). Tokens are normalized (lowercased, hyphens removed) and matched after NLTK lemmatization. 

\paragraph{2. LLM-based extraction.}
Each filtered sentence is sent to gpt-4o-mini with a fixed prompt that instructs the model to: (i)~extract descriptions of a person's physical attributes from the given text by outputing each description as a noun phrase that removes gender pronouns (e.g., ``his narrow jaw'' $\rightarrow$ ``a narrow jaw''); this requires both a body part and a characterizing modifier (e.g., ``dark brown hair'', ``piercing green eyes'') and to skip bare body-part mentions (e.g., ``a beard'') or purely evaluative phrases without a body part (e.g., ``a slender girl''); (ii)~for each extracted attribute, assign a \textbf{gender} label (\texttt{male}, \texttt{female}, \texttt{non-binary}, or \texttt{unknown}) and provide reasoning.


\paragraph{3. Postprocessing and filtering.}
The pipeline postprocesses the then normalizes gender pronouns in the attribute strings: possessive ``his''/``her'' are replaced by ``a''/``an''; pronouns such as ``he''/``she''/``him''/``herself''/``himself''/``hers'' are removed. Specifically, we also classify the extracted attributes from novels in two categories in terms of whether their formats are consistent with the LLM-generated or human-written physical attributes: \textbf{(a) in-domain-format attributes} are attributes that follow exactly the same pattern  \{[(a/an)] + ADJ + NOUN\} (e.g., ``a narrow jaw'', ``black hair'', ``a long beard'', ``a wide smile''); \textbf{(b) out-of-domain-format attributes} are physical attributes that do not exactly match the pattern, including: more than two content words (e.g. ``bright green eyes'', ``vast muscular arms''), different word order or POS (e.g. noun before adjective, or no clear single adjective + single noun), longer or more complex noun phrases.

\subsection{Prompt for LLM-Based Extraction}
\label{app:extraction_prompt}
\begin{lstlisting}[label={lst:prompt}]
You are an expert at physical attribute extraction from text. Your task is to extract physical attributes and gender of a person from a book.
You will be given unstructured text from a book that potentially describes physical features of a person. Ignore descriptions of clothes.
You should first extract all the descriptions of a person's physical attributes and convert them into a **noun phrase structure that removes gender pronouns like 'his' or 'her'**. Descriptions of a person's physical attributes are defined as adjectives or noun phrases describing the characteristics of a body part. You should extract complete noun phrases including both the body part and its description, such as "a narrow jaw", "dark brown hair", "piercing green eyes", and skip extraction of a body part without its description (e.g., "a beard"). A word or phrase merely referring to a body part without any chacracteristic description in the text, such as "a beard" or "his hair", should not be extracted; a mere description without a body part in the text also should not be extracted (e.g., "a slender girl"). If there is no description of physical attributes that meet the extraction criteria, simply return an empty list. 
Then, for each physical attribute, you should extract or infer the gender of the person described by the attribute, by labeling it as "male", "female", or "non-binary" if there is solid evidence in the text supporting one of them, or "unknown" if there is no evidence for inference.
Provide your reasoning and justification for each attribute's gender label.

For example:
<text>
Harry Potter was a young man with a thin, angular face, sharp features, and a narrow jaw.
</text>
[{"attribute":"a thin face", "reasoning":"The attribute is used to describe 'a young man', explicitly suggesting the person's gender; 'Harry' is also a common male name.", "gender":"male"}, {"attribute":"an angular face", "reasoning":"The attribute is used to describe 'a young man', explicitly suggesting the person's gender; 'Harry' is also a common male name.", "gender":"male",}, {"attribute":"sharp features", "reasoning":"The attribute is used to describe 'a young man', explicitly suggesting the person's gender; 'Harry' is also a common male name.", "gender":"male", }, {"attribute":"a narrow jaw", "reasoning":"The attribute is used to describe 'a young man', explicitly suggesting the person's gender; 'Harry' is also a common male name.", "gender":"male"}]

<text>
Her hair was dark brown and curly, and her eyes were a piercing green, staring at the man intently.
</text>
[{"attribute":"dark brown hair", "reasoning":"The text uses the pronoun 'her' to refer to the person with the current attribute ('dark brown hair'), explicitly suggesting the person's gender as female.", "gender":"female"}, {"attribute":"curly hair", "reasoning":"The text uses the pronoun 'her' to describe the person whose current attribute ('currly hair') is being described, explicitly suggesting the person's gender as female.", "gender":"female"}, {"attribute":"piercing green eyes", "reasoning":"The text uses the pronoun 'her' to describe the person whose eyes are being described, explicitly suggesting the person's gender as female.", "gender":"female"}]

<text>
A thick lump grew in his throat.
</text>
[]

<text>
The nurse's eyes were dark blue.
</text>
[{"attribute":"dark blue eyes", "reasoning":"The text describes the nurse's eyes as 'dark blue' without any clues about the person's gender.", "gender":"unknown"}]

<text>
His  foot  pushed  mine  off  the  gas  pedal.
</text>
[]

<text>
He took her small pink hand in his own frail spotted one and gave it a gentle squeeze.
</text>
[{"attribute":"a small pink hand", "reasoning":"The text uses the pronoun 'her' to refer to the person with the current attribute ('a small pink hand'), explicitly suggesting the person's gender as female.", "gender":"female"}, {"attribute":"a frail spotted hand", "reasoning":"The text uses the pronoun 'his' to refer to the person with the current attribute ('a frail spotted hand'), explicitly suggesting the person's gender as male.", "gender":"male"}]

<text>
Hers was not a pretty face, alas.
</text>
[{"attribute":"a pretty face", "reasoning":"The text uses the pronoun 'hers' to refer to the person with the current attribute ('a pretty face'), explicitly suggesting the person's gender as female.", "gender":"female"}]

\end{lstlisting}

\section{Zero-Shot LLM Inference Experiments}

\subsection{Prompt Template}
\label{app:model_exp_prompt}
\begin{lstlisting}[label={lst:prompt}]
How likely is it for someone to say a {person_term} has {attribute}? Your answer should only be a single number between 1 (Not at all) and 7 (Extremely likely).
Answer:

\end{lstlisting}

\subsection{Method Details}
\label{app:model_exp_method}

Two complementary strategies were adopted to elicit ratings based on varied constraints from different LLMs: \textbf{(1) direct generation}: we directly extract the verbalized ratings in the model-generated response; \textbf{(2) top-logit token extraction}: we examine the model's output distribution at the first generated token position: for open-weight models, we rank top k tokens with the highest logit values, select the token that contains any number in 1--7 can be extracted, and use the number as model rating. This is a more tolerant approach than simply selecting the highest-logit token among the seven tokens 1-7, as it considers common cases of zero-shot output that contain blanks adjacent to the actual answering, causing unexpected tokenization results; for API-based models, we similarly retrieve log-probabilities for the top-ranked tokens (if available) and identify the highest-probability token containing a valid rating. Incorporating both strategies ensures the collection of model rating preferences even without demanding their instruction-following abilities, and provides a robust and thorough investigation of models' encoding of gender in physical descriptions from not only the model's explicit textual response but also its underlying distributional preferences.

Method 1 and 2 showed a close-to-perfect agreement rate, serving as sanity check, except for the Mistral-7b models. Considering the higher accessibility of method 1, we report the ratings of the directly generated responses from models as the primary results.


\subsection{Complete Results}
\label{app:model_exp_results}

\autoref{fig:model_r_all} in the main body text presents the complete ratings in comparison with human ratings across genders for all models. Below, \autoref{tab:model_exp_all_results} shows the per-gender RMSE \& Pearson $r$ statistics of all evaluated models in the experiment of gender association on \dataset, with significance values included in the brackets.



\begin{table}[ht]
\centering
\scriptsize
\caption{Per-gender RMSE \& Pearson $r$ statistics of all evaluated models in the experiment of gender association on \dataset.}

\begin{tabular}{llcccccc}

\toprule

\multirow{2}{*}{\textbf{Model}} & \multirow{2}{*}{\textbf{Dataset}}
& \multicolumn{2}{c}{\textbf{woman}}
& \multicolumn{2}{c}{\textbf{man}}
& \multicolumn{2}{c}{\textbf{non-binary}} \\

\cmidrule(lr){3-4} \cmidrule(lr){5-6} \cmidrule(lr){7-8}

& & RMSE & Pearson $r$ ($p$)
& RMSE & Pearson $r$ ($p$)
& RMSE & Pearson $r$ ($p$) \\

\midrule

\multirow{4}{*}{Meta-Llama-3-8B}
& LLM-generated   & 2.67 & 0.60 (0.000) & 2.76 & 0.50 (0.001) & 2.95 & 0.16 (0.308) \\
& Novel-extracted & 2.82 & 0.56 (0.000) & 3.09 & 0.49 (0.000) & 2.85 & 0.19 (0.060) \\
& Human-written   & 2.57 & 0.59 (0.000) & 2.90 & 0.39 (0.005) & 3.06 & 0.23 (0.102) \\
& All             & 2.72 & 0.58 (0.000) & 2.97 & 0.45 (0.000) & 2.93 & 0.19 (0.011) \\

\midrule

\multirow{4}{*}{Llama-3-8B-Inst}
& LLM-generated   & 1.70 & 0.36 (0.019) & 2.00 & 0.47 (0.002) & 1.96 & 0.05 (0.775) \\
& Novel-extracted & 1.90 & 0.48 (0.000) & 2.21 & 0.48 (0.000) & 1.84 & 0.32 (0.002) \\
& Human-written   & 1.48 & 0.59 (0.000) & 2.00 & 0.38 (0.007) & 1.77 & 0.49 (0.000) \\
& All             & 1.75 & 0.49 (0.000) & 2.11 & 0.44 (0.000) & 1.85 & 0.33 (0.000) \\

\midrule

\multirow{4}{*}{Mistral-7B-v0.3}
& LLM-generated   & 3.13 & 0.42 (0.006) & 2.90 & 0.59 (0.000) & 2.11 & 0.19 (0.227) \\
& Novel-extracted & 2.75 & 0.36 (0.000) & 2.67 & 0.38 (0.000) & 1.76 & 0.34 (0.001) \\
& Human-written   & 2.46 & 0.63 (0.000) & 2.72 & 0.42 (0.002) & 1.40 & 0.42 (0.003) \\
& All             & 2.77 & 0.45 (0.000) & 2.73 & 0.42 (0.000) & 1.76 & 0.32 (0.000) \\

\midrule

\multirow{4}{*}{Mistral-7B-Inst}
& LLM-generated   & 1.43 & 0.58 (0.000) & 1.79 & 0.52 (0.000) & 1.85 & 0.22 (0.160) \\
& Novel-extracted & 1.89 & 0.61 (0.000) & 1.86 & 0.57 (0.000) & 1.87 & 0.32 (0.002) \\
& Human-written   & 1.72 & 0.69 (0.000) & 1.65 & 0.65 (0.000) & 1.69 & 0.46 (0.001) \\
& All             & 1.75 & 0.63 (0.000) & 1.79 & 0.58 (0.000) & 1.82 & 0.35 (0.000) \\

\midrule

\multirow{4}{*}{OLMo2-7B-1124}
& LLM-generated   & 1.49 & 0.72 (0.000) & 2.05 & 0.20 (0.199) & 2.92 & 0.25 (0.117) \\
& Novel-extracted & 1.90 & 0.62 (0.000) & 2.20 & 0.53 (0.000) & 2.70 & 0.35 (0.001) \\
& Human-written   & 1.71 & 0.68 (0.000) & 1.93 & 0.53 (0.000) & 2.82 & 0.29 (0.040) \\
& All             & 1.77 & 0.66 (0.000) & 2.10 & 0.47 (0.000) & 2.78 & 0.30 (0.000) \\

\midrule

\multirow{4}{*}{OLMo-2-7B-DPO}
& LLM-generated   & 1.37 & 0.36 (0.018) & 1.34 & 0.24 (0.128) & 1.66 & 0.25 (0.110) \\
& Novel-extracted & 1.41 & 0.56 (0.000) & 1.38 & 0.48 (0.000) & 1.71 & 0.36 (0.000) \\
& Human-written   & 1.34 & 0.64 (0.000) & 1.08 & 0.61 (0.000) & 1.69 & 0.29 (0.042) \\
& All             & 1.38 & 0.56 (0.000) & 1.30 & 0.47 (0.000) & 1.69 & 0.34 (0.000) \\

\midrule

\multirow{4}{*}{Qwen2.5-14B}
& LLM-generated   & 1.04 & 0.72 (0.000) & 1.44 & 0.60 (0.000) & 1.52 & 0.41 (0.007) \\
& Novel-extracted & 1.22 & 0.75 (0.000) & 1.60 & 0.59 (0.000) & 1.64 & 0.60 (0.000) \\
& Human-written   & 1.19 & 0.82 (0.000) & 1.32 & 0.80 (0.000) & 1.38 & 0.51 (0.000) \\
& All             & 1.17 & 0.76 (0.000) & 1.49 & 0.63 (0.000) & 1.55 & 0.56 (0.000) \\

\midrule

\multirow{4}{*}{Qwen2.5-14B-Inst}
& LLM-generated   & 0.99 & 0.68 (0.000) & 1.03 & 0.53 (0.000) & 0.64 & 0.26 (0.097) \\
& Novel-extracted & 1.17 & 0.72 (0.000) & 1.31 & 0.58 (0.000) & 0.87 & 0.61 (0.000) \\
& Human-written   & 0.92 & 0.78 (0.000) & 0.89 & 0.70 (0.000) & 0.67 & 0.40 (0.004) \\
& All             & 1.07 & 0.72 (0.000) & 1.15 & 0.57 (0.000) & 0.77 & 0.56 (0.000) \\

\midrule

\multirow{4}{*}{gpt-4o}
& LLM-generated   & 0.98 & 0.76 (0.000) & 1.08 & 0.59 (0.000) & 0.61 & 0.20 (0.211) \\
& Novel-extracted & 1.23 & 0.72 (0.000) & 1.36 & 0.69 (0.000) & 0.84 & 0.62 (0.000) \\
& Human-written   & 0.92 & 0.80 (0.000) & 1.01 & 0.65 (0.000) & 0.72 & 0.43 (0.002) \\
& All             & 1.10 & 0.74 (0.000) & 1.22 & 0.63 (0.000) & 0.76 & 0.57 (0.000) \\

\bottomrule

\end{tabular}

\label{tab:model_exp_all_results}

\end{table}

\section{Training Experiments for the Proxy Predictor Model}
\label{app:model_training}

\subsection{Training Details}

We use the LLM-generated subset of \dataset~ for training, validation, and in-domain evaluation (split ratio = 60/15/25), and reserve the novel-derived and human-written attribute sets as out-of-domain evaluation benchmarks. Complementing the zero-shot generation approach in \autoref{sec:model_exp}, we train each model with an added linear regression head to predict aggregated human ratings of physical attributes for different gender categories $c \in \{\textit{woman}, \textit{man}, \textit{non-binary person}\}$. For each training instance, the model takes as input a sentence describing a physical attribute with respect to a target gender (e.g., ``a woman has broad shoulders''). The mean-pooled final-layer hidden representation is passed through a one-dimensional linear projection to produce a scalar prediction corresponding to the human likelihood rating for that attribute–gender pair. This formulation treats the task as continuous regression, avoiding reliance on explicit numerical generation and providing a more direct probe of model–human alignment.

The model is optimized using a mean squared error (MSE) loss between predicted scores and averaged human ratings. Let $y_{ij}$ denote the rating provided by rater $j$ for item $i$, and $\bar{y}_i = \frac{1}{k}\sum_{j=1}^k y_{ij}$ the mean across $k$ raters. Training minimizes $\mathcal{L} = \frac{1}{N}\sum_{i=1}^N (\hat{y}_i - \bar{y}_i)^2$, where $\hat{y}_i$ is the model prediction and $N$ is the number of samples. Training on aggregated ratings encourages the model to capture the shared signal across annotators while attenuating idiosyncratic noise.

To ensure fair comparison across models with different architectures and scales, we adopt a standardized hyperparameter search template, with model-specific ranges for sensitive parameters such as learning rate and effective batch size. Models are compared based on their best performance across the search. Each model is allocated the same tuning budget and evaluated under consistent cross-validation and multi-seed settings to improve robustness. Full training details and configurations are provided in \autoref{app:model_training}.

\subsection{Full Configurations for Hyperparameter Search}
\label{app:hp_config}

\autoref{tab:hp_config} lists the hyperparameter configuration for hyperparameter search in the training ablation experiments, aiming for a relatively controlled comparison across
model family, scale, and instruction-tuning variants.

\begin{table*}[htbp]
\centering
\renewcommand{\arraystretch}{1}
\begin{tabular}{p{4cm} p{9cm}}
\toprule
\textbf{Models} &
Llama~3-8B (base and instruct), Mistral-7B (base and instruct), Qwen2.5-3B(base and instruct), Qwen2.5-7B(base and instruct), Qwen2.5-14B(base and instruct), Qwen2.5-32B(base and instruct), OLMo-2-7B (base, SFT, and DPO), and GPT-OSS-20B.
\\
\midrule
\textbf{Learning rate} &
$\{1\times10^{-5},\,5\times10^{-5},\,1\times10^{-4},\,2\times10^{-4},\,5\times10^{-4}\}$ \\
\textbf{Batch size} &
$\{2,\,4\}$ \\
\textbf{LoRA rank} ($r$) &
$\{8,\,16\}$ \\
\textbf{LoRA scaling} ($\alpha$) &
$\{16,\,32\}$ \\
\midrule
\textbf{Early stopping} &
Enabled (patience = 6) \\
\textbf{Cross-validation} &
K-fold ($K = 5$) \\
\textbf{Seeds/Replicates} & Aggregated from 3 independent runs per fold (seeds: 42, 123, 456) \\
\bottomrule
\end{tabular}
\caption{Training hyperparameter configuration, aiming for a relatively controlled comparison across model family, scale, and instruction-tuning variants.}
\label{tab:hp_config}
\end{table*}


\section{How to Interpret the Exploratory LitBank Analyses}
\label{app:novels}



It is important to note that the analyses in \autoref{subsec:novels} reflect predicted gender associations as they are held by the US participants we recruited in recent years. Since gender associations change over time, the following results should not be interpreted as a clear window into the authors' absolute biases decades ago. Instead, it's a window into the associations a common reader may take away today and comparative analyses between groups of authors can inform hypotheses about potentially persistent gendered perceptions.

\end{document}